\documentclass[10pt,journal,compsoc]{IEEEtran}
\ifCLASSOPTIONcompsoc
  \usepackage[nocompress]{cite}
\else
  \usepackage{cite}
\fi
\usepackage{microtype}
\usepackage{wrapfig}
\usepackage{stfloats}
\usepackage{booktabs}
\usepackage{amstext,amsmath,amssymb}
\allowdisplaybreaks[3] 

\usepackage[utf8]{inputenc} 
\usepackage[T1]{fontenc}    
\usepackage{hyperref}       
\hypersetup{
	colorlinks=true,
	linkcolor=black,
	anchorcolor=black,
    citecolor=black,
    urlcolor=black
}
\usepackage{url}            
\usepackage{booktabs}       
\usepackage{amsfonts}       
\usepackage{nicefrac}       
\usepackage{microtype}      
\usepackage{xcolor}         
\usepackage{color}
\usepackage{colortbl}
\definecolor{TODO}{RGB}{238,34,12}
\definecolor{REVISE}{RGB}{238,34,12}
\usepackage{caption}
\usepackage[font=small]{subcaption}

\usepackage{algorithm}
\usepackage{algorithmic}
\usepackage{multirow}
\usepackage{enumitem}
\usepackage[normalem]{ulem}
\usepackage{ragged2e}

\usepackage{marvosym}
\newcommand{\gr}{\rowcolor[gray]{.95}}

\ifCLASSINFOpdf
   \usepackage[pdftex]{graphicx}
\else
\fi
\begin{document}
%
\title{MetaVIM: Meta Variationally Intrinsic Motivated Reinforcement Learning 
for Decentralized \\Traffic Signal Control}
%
%
%
%

\author{Liwen~Zhu, Peixi~Peng, Zongqing~Lu, Yonghong~Tian~\IEEEmembership{Fellow,~IEEE} 
\IEEEcompsocitemizethanks{
\IEEEcompsocthanksitem Corresponding authors: Peixi Peng and Yonghong Tian. \\
\IEEEcompsocthanksitem Liwen~Zhu, Peixi~Peng, Zongqing~Lu and Yonghong~Tian are with National Engineering Research Center of Visual Technology, School of Computer Science, Peking University, China, and are also with Pengcheng Laboratory, Shenzhen, China. E-mail:\{liwenzhu,pxpeng,zongqing.lu,yhtian\}@pku.edu.cn.
\IEEEcompsocthanksitem This work is partially supported by grants from the Key-Area Research and Development Program of Guangdong Province under contact No. 2020B0101380001 and grants from the National Natural Science Foundation of China under contract No. 62027804, No. 61825101 and No. 62088102. The
computing resources of Pengcheng Cloudbrain are used in this research.
\protect\\
}
}

%
%

\markboth{IEEE Transactions on Knowledge and Data Engineering}{}
%




\IEEEtitleabstractindextext{%
\justifying  
\begin{abstract}
Traffic signal control aims to coordinate traffic signals across intersections to improve the traffic efficiency of a district or a city. Deep reinforcement learning (RL) has been applied to traffic signal control recently and demonstrated promising performance where each traffic signal is regarded as an agent. 
However, there are still several challenges that may limit its large-scale application in the real world. 
On the one hand, the policy of the current traffic signal is often heavily influenced by its neighbor agents, and the coordination between the agent and its neighbors needs to be considered. Hence, the control of a road network composed of multiple traffic signals is naturally modeled as a multi-agent system, and all agents' policies need to be optimized simultaneously. On the other hand, once the policy function is conditioned on not only the current agent's observation but also the neighbors', the policy function would be closely related to the training scenario and cause poor generalizability because the agents in various scenarios often have heterogeneous neighbors. 
To make the policy learned from a training scenario generalizable to new unseen scenarios, a novel Meta Variationally Intrinsic Motivated (MetaVIM) RL method is proposed to learn the decentralized policy for each intersection that considers neighbor information in a latent way. Specifically, we formulate the policy learning as a meta-learning problem over a set of related tasks, where each task corresponds to traffic signal control at an intersection whose neighbors are regarded as the unobserved part of the state. Then, a learned latent variable is introduced to represent the task's specific information and is further brought into the policy for learning. In addition,  to make the policy learning stable, a novel intrinsic reward is designed to encourage each agent's received rewards and observation transition to be predictable only conditioned on its own history. Extensive experiments conducted on CityFlow demonstrate that the proposed method substantially outperforms existing approaches and shows superior generalizability.
\end{abstract}

}

\maketitle

\IEEEdisplaynontitleabstractindextext

%
\IEEEpeerreviewmaketitle

\IEEEraisesectionheading{\section{Introduction}\label{sec:introduction}}

%
%
%
%

 

\IEEEPARstart{T}{raffic} signals that direct traffic movements play an important role in efficient transportation. Most conventional methods aim to control traffic signals by fixed-time plans \cite{koonce2008traffic} or hand-crafted heuristics \cite{varaiya2013max}. However, the pre-defined rules cannot adapt to dynamic and uncertain traffic conditions. Recently, deep reinforcement learning (RL) \cite{guo2021urban,jintao2020learning,pan2020spatio,he2020spatio,tong2021combinatorial,wang2020deep,gu2020exploiting,liu2021urban,zhang2021periodic} has been applied in Intelligent Transportation Systems (ITS) and demonstrated promising performance. Specifically, recent works \cite{wei2018intellilight,wei2019presslight,zang2020metalight} use RL for traffic signal control, and provide a promising way to handle dynamic and uncertain traffic conditions, where a RL agent employs a deep neural network to control an intersection and the network is learned by directly interacting with the environment.

The most straightforward RL baseline considers each intersection independently and models the task as a single agent RL problem \cite{wei2018intellilight}. However, the observation, received reward and dynamics of each traffic signal are closely related to its neighbors, and the coordination between signals should be modeled. Hence, optimizing traffic signal control in a large-scale road network could be modeled as a multi-agent reinforcement learning (MARL) problem. Prior works \cite{kuyer2008multiagent,van2016coordinated} in this domain resort to centralized training to ensure that agents learn to coordinate, they demonstrate that communication among agents could help coordination. However, as the joint action space grows exponentially with the number of agents, it is infeasible or costly in realistic deployment, and training emergent communication protocols also remains a challenging problem. 
In addition, once the policy function is conditioned on not only the current agent's observation but also the neighbors', the policy function would be closely related to the training scenario and cause poor generalizability because the agents in various scenarios often have heterogeneous neighbors. Therefore, learning decentralized policies opens up a window of new opportunities for advanced MARL research in this area.


\begin{figure}[t]
\centering
\includegraphics[width=0.49\textwidth]{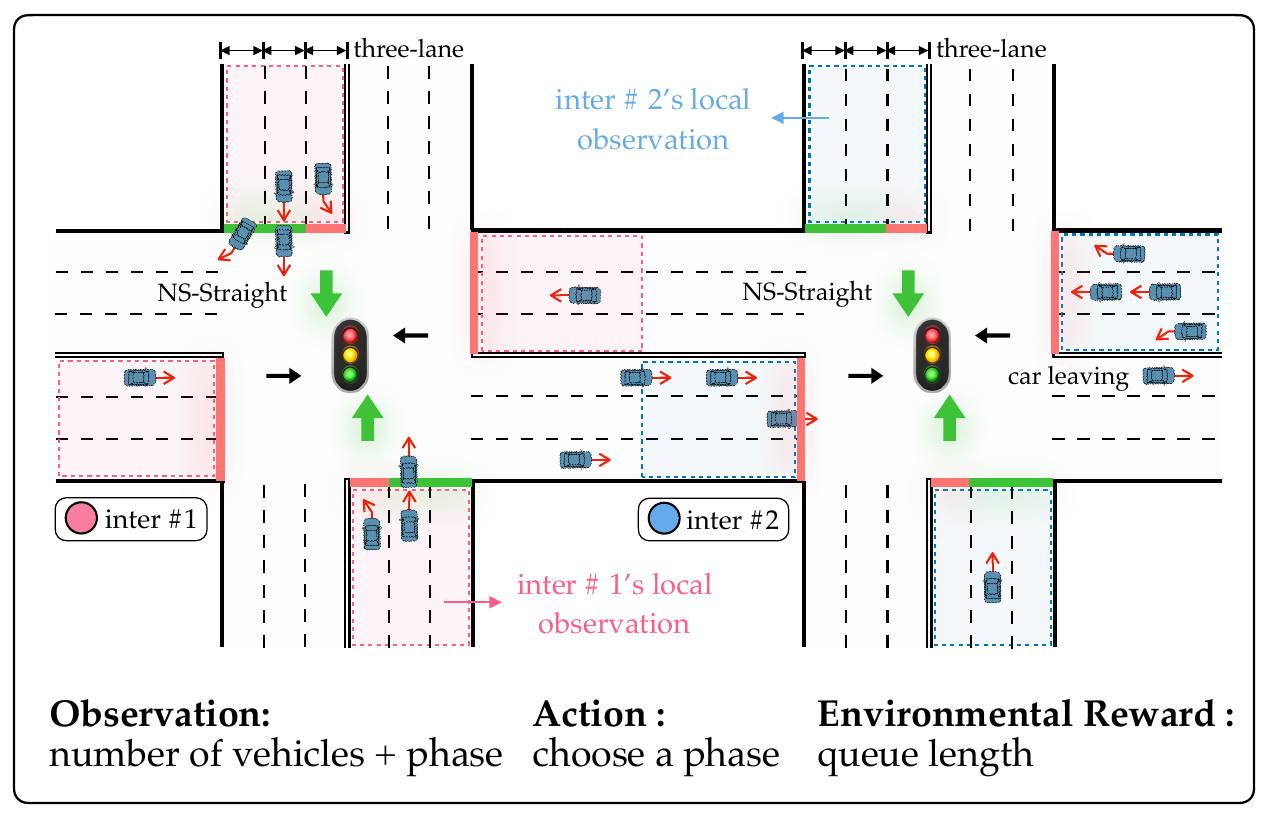}
 		\vspace{-0.3cm}
 		\caption{Illustration of the multi-intersection control scenario where policies and rewards are affected by neighboring policies, i.e., the instability of the multi-agent scenario.}
 		\label{fig:new_scenario}
 		\vspace{-0.5cm}
\end{figure}

To learn effective decentralized policies, there are two main challenges. Firstly, it is impractical to learn an individual policy for each intersection in a city or a district containing thousands of intersections. Parameter sharing may help. However, each intersection has a different traffic pattern, and a simple shared policy hardly learns and acts optimally at all intersections. To handle this challenge, we formulate the policy learning in a road network as a meta-learning problem, where traffic signal control at each intersection corresponds to a task, and a policy is learned to adapt to various tasks. Reward function and state transition of these tasks vary but share similarities since they follow the same traffic rules and have similar optimization goals. Therefore, we represent each task as a learned and low-dimensional latent variable obtained by encoding the past trajectory in each task. The latent variable is a part of the input of the policy, which captures task-specific information and helps improve the policy adaption. 

Secondly, even for a specific task, the received rewards and observations are uncertain to the agent, as illustrated in Fig.~\ref{fig:new_scenario}, which make the policy learning unstable and non-convergent. Even if the agent performs the same action on the same observation at different timesteps, the agent may receive different rewards and observation transitions because of neighbor agents' different actions. In this case, the received rewards and observation transitions of the current agent could not be well predicted only conditioned on its own or partial neighbors' observations and performed actions. To avoid this situation, four decoders are introduced to predict the next observations and rewards without neighbor agents' policies or with partially neighbor agents, respectively. In addition, an intrinsic reward is designed to reduce the bias among different predictions and enhance learning stability. In other words, the design of the decoders and intrinsic reward is similar to the law of contra-positive. The unstable learning will cause the predicted rewards and observation transitions unstable in a decentralized way, while our decoders and intrinsic reward encourage the prediction convergent. In addition, from the perspective of information theory, the intrinsic reward design makes the policy of each agent robust to neighbours' polices, which could make the learned policy easy to transfer. 

Intrinsic motivation refers to reward functions that allow agents to learn useful behaviors across various tasks. Previous approaches to intrinsic motivation often focus on curiosity \cite{pathak2017curiosity}, imagination \cite{andrychowicz2017hindsight} or synergy \cite{chitnis2019intrinsic}, these approaches rely on hand-crafted rewards specific to the environment, or limit to the bimanual manipulation tasks. Such structural constraints make it impossible to achieve independent training of MARL agents across multiple environments. Here, we consider the problem of deriving intrinsic motivation as an exploration bias from other agents in MARL. Our key idea is that a good guiding principle for intrinsic motivation is to make approximations robust to a dynamic environment. Intrinsic motivation is assessed using comparison reasoning: at each timestep, an agent measures the effect of another agent's policy on its model's predictions. Neighbor's policies that lead to a relatively higher change in an agent's own model estimates are considered highly influential, and because the explicit effect of neighbor's influence in this task is to increase incoming traffic volumes, which exacerbates the agent's own traffic load. Therefore, we minimize this influence bias to make the model's estimations free from interference from others. We show that this inductive bias will drive agents to learn effective decentralized policies.

We conduct extensive experiments on CityFlow \cite{zhang2019cityflow} in public datasets Hangzhou (China), Jinan (China), New York (USA), and our derived dataset Shenzhen (China) road networks under various traffic patterns, and empirically demonstrate that our proposed method can achieve state-of-the-art performances over the above scenarios. Furthermore, our method shows superior adaptivity in transfer experiments. 
In summary, the main contributions of this paper are concluded as three aspects:

1) The traffic signal control  is modeled as a meta-learning problem over a set of related tasks, and a novel method MetaVIM is proposed to learn decentralized policy to handle large-scale traffic lights.

2) A learnable latent variable is introduced to represent task-specific information by performing approximate inference on the task, and make the policy function shareable across the tasks.

3) A novel intrinsic reward is designed to tackle the challenge of unstable policy learning in dynamically changing traffic environments. We show that this design will drive agents to learn effective decentralized policies.


The paper is structured as follows. We describe the related work in Section~\ref{sec:related_work}, and the MARL setting in Section~\ref{sec:problem_statement}. Section~\ref{sec:method} introduces the details of the proposed method. Section~\ref{sec:experiments} presents experimental results that empirically demonstrate the efficacy of MetaVIM. Finally, conclusions and future work are discussed in Section~\ref{sec:conclusions}.

\section{Related Work}\label{sec:related_work}
\subsection{Conventional and Adaptive Traffic Signal Control} \label{sec:related_work_conventional_tsc}


Most conventional traffic signal control methods are designed based on fixed-time signal control \cite{webster1958traffic}, actuated control \cite{chiu1992adaptive} or self-organizing traffic signal control \cite{cools2013self}. These approaches rely on expert knowledge and often perform unsatisfactorily in complicated real-world situations. To solve this problem, several optimization-based methods have been proposed to optimize average travel time, throughput, \textit{etc.}, which decide the traffic signal plans according to the dynamical observed data. 
Specifically,  the method \cite{varaiya2013max} calculates the difference between the number of upstream and downstream vehicles and then sorts the values to determine the phase order. 
Besides, several methods \cite{gao2016optimizing,gao2018solving,celtek2020real,cheng2020monte} consider the scheduling of urban traffic light as the model-based optimization problem and employ search based methods to solve it. Furthermore, the dynamic programming \cite{chen2016improved}, mixed-integer linear programming \cite{he2016traffic} and non-linear programming models \cite{mohebifard2019optimal} are also used. Please see \cite{qadri2020state} for more details. Although effective in some situations, these approaches typically rely on prior assumptions which may fail in some cases, or require environment model which is unavailable and unreliable in complex scenarios.

\subsection{RL-based Traffic Signal Control} 

RL-based traffic signal control methods aim to learn the policy from interactions with the environment. Earlier studies use tabular Q-learning \cite{el2013multiagent,abdoos2013holonic,dusparic2009distributed,abdoos2011traffic} where the states are required to be discretized and low-dimensional. To handle  more complex continuous state, recent advances employ deep RL  to map the high-dimensional state representations (such as images or feature vectors) into actions.

Efforts have been made to design strategies that formulate the task as a single agent \cite{wei2018intellilight,huang2021modellight,zang2020metalight,jiang2021dynamic} or some isolated intersections \cite{zheng2019diagnosing,zheng2019learning,xiong2019learning,chen2020toward,oroojlooy2020attendlight,zhang2020planlight}
, i.e., each agent makes decision for its own. This type of methods is usually easy to scale, but may have difficulty to achieve global optimal performance due to the lack of collaboration. To address the problem, another way is to  jointly model the action among learning agents with centralized optimization \cite{van2016coordinated,kuyer2008multiagent}. However, 
as the number of agents increases, joint optimization usually leads to  dimensional explosion, which has inhibited the widespread adoption of such methods to a large-scale traffic signal control. To overcome the difficulty, another type of methods are implemented in a decentralized manner. For example, the methods proposed in \cite{el2013multiagent,chu2019multi} directly add neighboring information into states, and the neighbors' hidden features are merged into states in \cite{nishi2018traffic,wei2019colight,yu2020macar,guo2021urban}. Compared with them, our method uses neighbor information to form intrinsic motivation rather than as additional input of the policy. It makes our method easy to transfer to a new scenario which may have different neighbour topology with the training scenario. Besides, the neighborhood travel time is optimized in \cite{xu2021hierarchically} as an additional reward. However, simple concatenation of neighboring information is not reasonable enough because the influence of neighboring intersections is not balanced.

To make the policy transferable, traffic signal control is also modeled as a meta-learning problem in \cite{zang2020metalight,zhang2020generalight,huang2021modellight}. Specifically, the method in \cite{zang2020metalight} performs meta-learning on multiple independent MDPs and ignores the influences of neighbor agents. A data augmentation method is proposed in \cite{zhang2020generalight} to generates diverse traffic flows to enhance meta-RL, and also regards agents as independent individuals, without explicitly considering neighbors. In addition, a model-based RL method is proposed in \cite{huang2021modellight} for high data efficiency. However it may introduce cumulative errors due to error of the learned environment model and it is hard to achieve the asymptotic performance of model-free methods. Our method  both belongs to meta-RL paradigms, the main advantages are  two main aspects Firstly, we consider the neighbour information during the meta-learning, which is critical for the multi-agent coordination. Secondly, our method learns a latent variable to represent task-specific information, which can not only balance exploration and exploitation \cite{zintgraf2019varibad}, but also help to learn the shared structures of reward and transition across tasks. As far as we know, our work is the first to propose an intrinsic motivation to enhance the robustness of the policy on traffic signal control. See Appendix \ref{sec:rl_based_tsc} for a brief overview of the above methods.


\subsection{Intrinsic Reward} \label{sec:related_work_intrinsic_reward}
Intrinsic motivation methods have been widely studied in the literature, such as handling  the difficult-to-learn dilemma in sparse reward environments \cite{yang2021ciexplore} or trading off the exploration and exploitation in non-sparse reward scenarios \cite{zintgraf2019varibad}. Most of the intrinsic reward approaches can be classified into two classes. The first class is counted-based paradigm, where agents are incentivized to reach infrequently visited states by maintaining state visitation counts \cite{bellemare2016unifying,tang2017exploration} or density estimators \cite{ostrovski2017count,burda2018large}. However, this paradigm is challenging in continuous or high-dimensional state space. The second is curiosity-based paradigm, in which agents are rewarded for high prediction error in a learned reward \cite{achiam2017surprise,pathak2017curiosity} or inverse dynamics model \cite{burda2018large,haber2018learning}. The uncertainty of the agent's assessment of its behavior can be measured as a curiosity for environmental exploration. 

Besides the above two classes, other intrinsic reward methods are mainly task-oriented and for a specific purpose. For example, the method in \cite{chitnis2019intrinsic} uses the discrepancy between the marginal policy and the conditional policy as the intrinsic reward for encouraging agents to have a greater social impact on others. The errors between the joint cooperative behaviors and the individual actions  are defined in \cite{jaques2019social} as an intrinsic reward, which is suitable for agent-pair tasks that rely heavily on collaboration, such as dual-arm robot tasks. Similar with them, the proposed intrinsic reward is specially designed
 for the traffic signal control task, and we adopt the discrepancy with/without neighbor agent's policies as a motivation for decentralized control in dynamically changing multi-agent scenarios, making the policy robust to the environment.

\subsection{Latent Variable in RL} \label{sec:realted_work_latent_variable}

A number of prior works have explored how RL can be cast in the framework of variational inference. Latent variable could transform the dynamically updated task-related information such as trajectories into a continuous lower-dimensional space. For example, \cite{zintgraf2021exploration} shows that exploring in latent space can enhance the representation of the environment. In meta-RL, the agent is not given prepared task-specific data to adapt to, and it must fully explore the environment to collect useful information. Privileged information generated during the learning phase can be used during meta-training to guide exploration, such as expert trajectories \cite{dorfman2020offline}, dense reward for meta-training but not testing \cite{rakelly2019efficient}, or ground-truth task IDs / descriptions \cite{kamienny2020learning,zhang2021metacure}. Similar work \cite{gupta2018meta} introduces temporally-extended exploration with latent variable. A series of methods \cite{zintgraf2019varibad,rakelly2019efficient,zintgraf2021exploration} use variational auto-encoders (VAE)\cite{kingma2013auto} structure to help explore the environment. A branch of context-based methods automatically learns to trade-off exploration and exploitation by maximizing average adaptation performance \cite{duan2016rl,zintgraf2019varibad}. 
Differently, we learn a dynamical latent variable for each task to present task-specific information and indicate the correspond agent's belief.

\section{Problem Statement}
\label{sec:problem_statement}
\begin{figure}[t]
 		\centering
 		\includegraphics[width=0.47\textwidth]{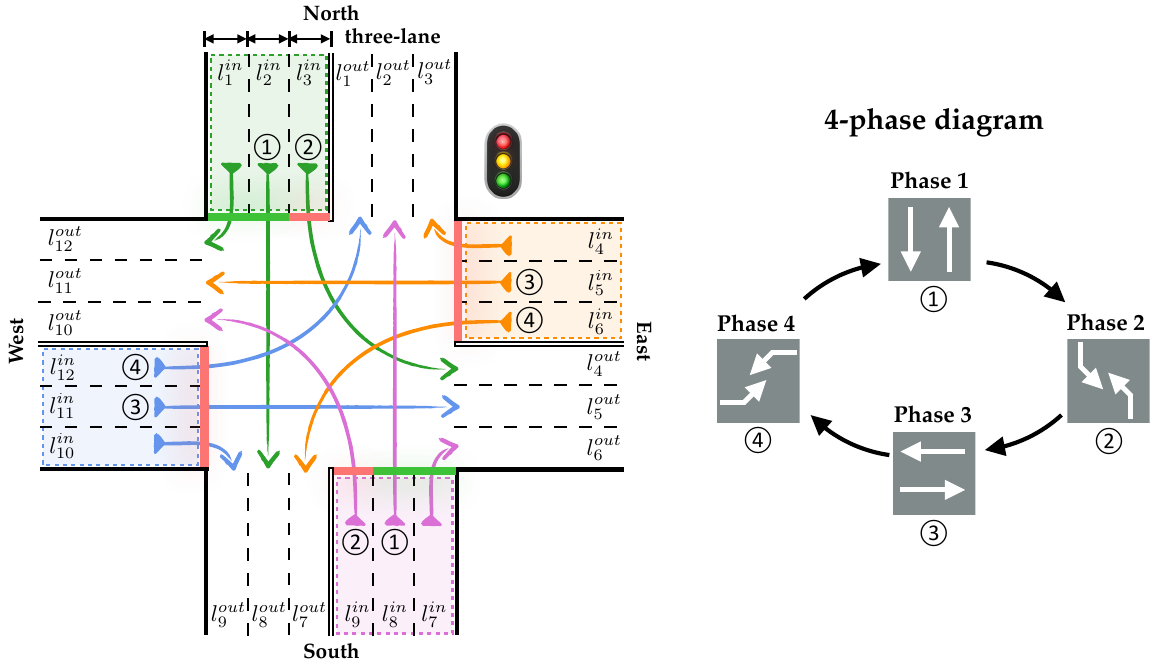} 
 		\vspace*{-0.3cm}
 		\caption{Illustration of incoming, outgoing lanes and 4-phase diagram.}
 		\label{fig:fig_lane}
 		\vspace*{-0.4cm}
\end{figure}

\subsection{Preliminary}\label{sec:preliminary}

In this paper, we investigate traffic signal control of multi-intersection as illustrated in Fig. \ref{fig:road_network}. In order to explain the basic concepts, we use a 4-way intersection as an example, depicted in Fig. \ref{fig:fig_lane}. Note that the concepts are easily generalized to different intersection structures (e.g., different numbers of entering approaches or phases).

\textbf{Definition 1} (\textit{Incoming/Outgoing Lanes})
For an intersection, the incoming lanes refer to the lanes where the vehicles are about to enter the intersection. In real world, most intersections are equipped with 4-way entering approaches, but some are 3-way or 5-way intersections. A standard 4-way intersection is shown in Fig. \ref{fig:fig_lane}, which consists of four approaches, i.e., "east", "south", "west" and "north". Each approach consists of three types of lanes, representing "left-turn", "straight" and "right-turn" directions from inner to outer. The outgoing lanes refer to the lanes where the vehicles are about to leave the intersection. Notes that vehicles on the incoming lanes are affected directly by the traffic signal at the current intersection. Therefore, we adopt the traffic information on the incoming lanes as part of the observation, which is the same as most existing works \cite{wei2019colight,wei2019presslight,chen2020toward,zang2020metalight}.

\textbf{Definition 2} (\textit{Phase})
Phase is a controller timing unit associated with the control of one or more movements, representing the permutation and combination of different traffic flows. At each phase, vehicles in the specific lanes can continue to drive. The 4-phase setting is the most common configuration in reality, but the number of phases can vary due to different intersection topologies (3-way, 5-way intersections, etc.). Fig. \ref{fig:fig_lane} 
illustrates a standard 4-phase setting: "north-south-straight", "north-south-left", "east-west-straight" and "east-west-left", "north-south-straight" means that the signal on the corresponding lanes are green. Note that the signal on the right-turn lanes is always green for consistency with real world.

\textbf{Definition 3} (\textit{Average Travel Time})
The travel time of a vehicle is the time discrepancy between entering and leaving a particular area. A vehicle from the origin to the destination (OD) is regarded as a travel. Average travel time of all vehicles in a road network is the most frequently used measure to evaluate the performance of traffic signal control \cite{xu2021hierarchically,zang2020metalight,chen2020toward,wei2019presslight,wei2019colight,wei2018intellilight,zhang2020planlight,zhang2020generalight,huang2021modellight}.

\subsection{Problem Definition}
The traffic signal control in a road network is modeled as a multi-agent system, where each signal is controlled by an agent. 
Before formulating the problem, we firstly design the learning paradigm by analyzing the characteristics of the traffic signal control (TSC). Due to the coordination among different signals, the most direct paradigm may be centralized learning. However, the global state information in TSC is not only highly redundant and difficult to obtain in realistic deployment, but also likely suffers from dimensional explosion. Moreover, once the policy function relies on the global state information or neighbors on execution, it is hard to transfer the policy from the training scenario to other unseen scenarios containing different road networks. Hence, it is natural to resort to the decentralized policy, which controls each signal only conditioned on its own history. However, the fully decentralized learning  ignores the coordination. If agents are behaved independently,  agents maximize their own rewards and may sacrifice the interests of others, it is difficult for the entire system to reach the optimum. Therefore, we model the task as Decentralized Partially Observable Markov Decision Process (Dec-POMDP) \cite{oliehoek2016concise}. The neighbors' information is considered, all agents' policies are optimized synchronously in training, while only the agent's observation history is used in the execution. 

Specifically, a Dec-POMDP could be denoted by a tuple $\mathcal{G}=<\mathcal{S}, \mathcal{N}, \mathcal{A}, \mathcal{O}, \mathcal{Z}, \mathcal{P}, \mathcal{R}, \mathcal{H}, \gamma>$. Each intersection in the scenario is controlled by an agent. $\mathcal{S}$ is the state space, and $s_t \in \mathcal{S}$ denotes the state at  time step $t$. Since the environment is partially observed, each agent $i$ only has access to its local observation $o_{i,t} \in \mathcal{O}$ that is acquired through an observation function $\mathcal{Z}(s): \mathcal{S}\rightarrow \mathcal{O}$. At current time, all agents perform joint action $ \boldsymbol{a}=\{a_i, \ldots, a_N\}, a_i\in \mathcal{A}$ and cause the state transition dynamics $\mathcal{P}\left(s_{t+1}|s_t, \boldsymbol{a}\right):=\mathcal{S} \times \mathcal{A}^{N} \rightarrow \mathcal{S}$ where $s_{t+1}$ is the next state. 
The observation-action history of agent $i$  at time $t$ is denoted as $\tau_{i,:t}$. $\mathcal{R}=\{\mathcal{R}_i\}_{i=1}^N$ is the reward for each agent. As stated in Sec. \ref{sec:agent_design}, the reward is calculated by the partial observation (queue length) and the observation transition may be  unstable in a multi-agent system. That is, even if the agent performs the same action on the same observation at different timesteps, the agent may receive different observation transitions because neighbor agents may perform different actions. Hence, we define the reward function of each agent as  $ r_{i,t}=\mathcal{R}_i(o_{i,t}, a_i,o_{i,t+1})$.
Our goal is to learn the decentralized policy $\pi_{i}\left(a_i \mid o_{i,t}\right)$ to maximize its cumulative rewards: $\sum_{t}{\gamma^t r_{i,t}}$ where $\gamma$ is the discounted factor. In the traffic signal task, the next observation of each agent not only relies on current agent's observation and performed action, but also is associated with neighbors' actions. Therefore, there exists an observation transition function $\mathcal{T}_i \big(o_{i,t+1} \mid o_{i,t}, a_{i}, \mathbf{a}^{-i}\big)$ for each agent $i$ where $\mathbf{a}^{-i}$ is the joint action of neighbors. 

Based on the above formulation, there are two key issues that need to be considered:
\begin{itemize}
    \item Since training policy for each realistic scenario is time-consuming even impossible, hence our goal is to learn the decentralized policy $\pi_{i}\left(a_i \mid o_{i,t}\right)$ from the given training scenario (i.e., a road network), which can be generalized to unseen scenarios. Thus the meta-learning framework is employed where the policy learning of each agent corresponds to a task. The reward function $R_i$ and observation transition $\mathcal{T}_i$ of these tasks vary but also share similarities since they follow the same traffic rules and have similar optimization goals. Therefore, we represent each task as a learned and low dimensional latent variable $m_i$. By incorporating $m_i$, we assume the reward, observation transition and policy functions could be shareable across tasks:
\begin{align}
  \mathcal{T}_i\big(o_{i,t+1} \mid o_{i,t}, a_i, \mathbf{a}^{-i}\big) & \approx \mathcal{T}\big(o_{i,t+1} \mid o_{i,t}, a_i, \mathbf{a}^{-i}, m_i\big),\nonumber \\
  \mathcal{R}_i\left(o_{i,t}, a_i,o_{i,t+1}\right) & \approx \mathcal{R}\left(o_{i,t}, a_i,o_{i,t+1}, m_i\right),\nonumber \\
  \pi_{i}\left(a_i \mid o_{i,t}\right) & \approx \pi\left(a_i \mid o_{i,t},m_i \right),
\label{eq:framework}
\end{align}
which make the meta-learning possible.
    \item Since $\mathcal{R}_i$ not only rely on $o_{i,t}$ and $a_i$ but also $o_{i,t+1}$ which is generated by $\mathcal{T}_i$ and related with $a^{-i}$, hence learning  $\pi_{i}$ from $\mathcal{R}_i$ 
     may cause learning non-stationary because the agent may receive different rewards and observation transitions for the same action at the same observation. In this case, the received rewards and observation transitions of the current agent could not be well predicted only conditioned on its own observations and performed actions. Conversely, to avoid suffering such non-stationary, we hope the learned decentralized policy could make the observation transition and reward predictable. That is, based on the learned $\pi(a_i|o_{i,t})$, 
    there exists functions $\mathcal{T}_i^{\prime}$ and $\mathcal{R}_i^{\prime}$ satisfy that:
\begin{align}
    \mathcal{T}_i^{\prime}(o_{i,t+1} \mid o_{i,t}, a_i)& \approx \mathcal{T}_i\big(o_{i,t+1} \mid o_{i,t}, a_i, \mathbf{a}^{-i}\big),\nonumber \\
    \mathcal{R}_i^{\prime}(o_{i,t+1}, o_{i,t}, a_i)& \approx \mathcal{R}_i\big(o_{i,t+1}, o_{i,t}, a_i, \mathbf{a}^{-i} \big),\nonumber \\
    where~a_i &\sim \pi(a_{i}|o_{i,t}).
    \label{eq:goal}
\end{align}
\end{itemize}

\subsection{Agent Design} \label{sec:agent_design}

Each intersection in the system is controlled by an agent, and  here we present the detailed definitions of the agent. 

\begin{itemize}[leftmargin=15pt]
\item 

\textbf{Observation.} 
Each agent has its own local observation, including the number of vehicles on each incoming lane and the current phase of the intersection, where phase is the part of the signal cycle allocated to any combination of traffic movements, as explained in Section \ref{sec:preliminary}. Observation of agent $i$ is defined by
\begin{align}
    o_{i,t} = \{\mathcal{V}_{1}, \mathcal{V}_{2}, \ldots, \mathcal{V}_{M}, \mathtt{p}\},
\end{align}
where $M$ is the total number of incoming lanes and $\mathcal{V}_{M}$ means the number of vehicles in the $M^{th}$ incoming lanes, $\mathtt{p}$ is the current phase and represented as a one-hot vector. 

\item 
\textbf{Action.} At time $t$, each agent $i$ chooses a phase $\mathtt{p}$ as its action $a_i$, indicating the traffic signal should be set to phase $\mathtt{p}$. Note that the phases may organize in a sequential way in reality, while directly selecting a phase makes the traffic control plan more flexible. In a grid road network, each agent has four phases as described in Section \ref{sec:preliminary}. For a complex and  heterogeneous road network, we unite the phases of all structures together as the full action space, and mask the unavailable phases of each traffic signal as unavailable actions of the agent. 

\item 
\textbf{Reward.} We define the reward for agent $i$ as the negative of the queue length on incoming lanes. Note that optimizing queue length has been proved to be equivalent to optimizing average travel time in \cite{zheng2019diagnosing} under certain assumptions. Average travel time is a global criteria which cannot be optimized directly by an agent. Hence, queue length is widely used as reward in traffic signal control. Reward of agent $i$ is defined by
\begin{align}
    r_{i} =  -\sum^{M}_{m} q_{m},
\end{align}
where $q_{m}$ is the queue length on the incoming lane ${m}$, and 
$M$ is the total number of incoming lanes. 
\end{itemize}

\begin{figure*}[t]
 		\centering
 		\includegraphics[width=0.75\textwidth]{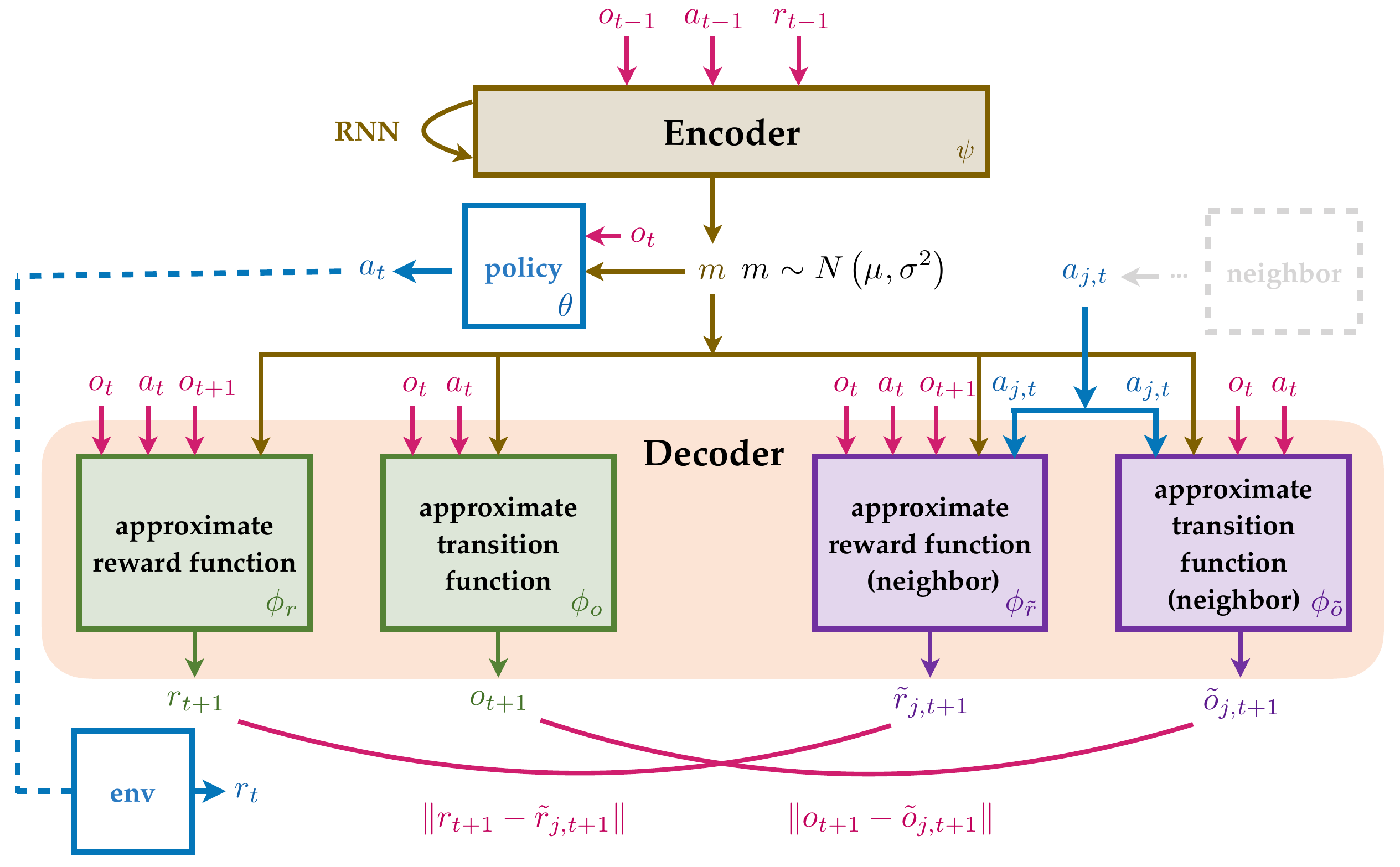} 
 		\caption{MetaVIM consists of a mVAE and a policy network. RL agent is augmented with a latent variable $m $, obtained by encoding over past trajectories $ \tau_{ :t}$ to represent the task-specific information.}
 		\label{fig:framework}
 		\vspace*{-0.3cm}
\end{figure*} 

\vspace{-3mm}
\section{Method}\label{sec:method}

In this section, we propose Meta Variationally Intrinsic Motivated (MetaVIM) method to achieve Eq. \ref{eq:framework} and Eq. \ref{eq:goal}, as illustrated in Fig.~\ref{fig:framework}. MetaVIM employs latent variable to represent each task to make the reward, observation transition and policy functions shareable. At the same time, MetaVIM makes the approximations and policy robust to neighbors by intrinsic reward design. From now on, we drop the sub- and superscript $i$ to denote the current agent for ease of notation.
We start by describing the overall architecture in Sec. \ref{sec:model}, and then elaborate the policies with latent variable in Sec. \ref{sec:policies_with_latent_variables} and intrinsic reward design in Sec. \ref{sec:intrinsic_reward}.
\vspace{-3mm}
\subsection{Model}\label{sec:model}
MetaVIM consists of a multi-head VAE (mVAE) and a policy network, where the mVAE consists of an encoder and four decoders:

\begin{itemize}
\item

\textbf{Encoder.} The encoder takes the past trajectories $\tau_{:t}$ up until  $t$ as input to calculate the latent variable $m$ in Eq.~\ref{eq:framework}, parameterised by $\psi$. The encoder could be:
\begin{align}
    \mathbf{e}^{\psi}\left(m \mid \tau_{:t} \right).
\end{align}
\item

\textbf{Decoder.} Four decoders are used to predict the environmental dynamics, and parameterised by $\phi_{\rm r},  \phi_{\rm \tilde{r}}, \phi_{\rm o}, \phi_{\rm \tilde{o}}$ respectively, where the former two are used to predict next reward with/without neighbors and the latter two are used to predict next observation with/without neighbors:
\begin{align}
\label{eq:decoder}
    &\mathbf{p}^{\phi_{\rm r}}\left(r_{t+1} \mid o_{t+1}, o_{t}, a, m\right), \\
    &\mathbf{p}^{\phi_{\rm o}}\left(o_{t+1} \mid o_{t}, a, m\right), \\
    &\mathbf{p}^{\phi \tilde{\rm r}}\left(\tilde{r}_{t+1} \mid o_{t+1}, o_{t}, a, a_{j}, m\right), \\
    &\mathbf{p}^{\phi_{\tilde{\rm o}}}\left(\tilde{o}_{t+1} \mid o_{t}, a, a_{j}, m\right),
\end{align}
where $j$ is any neighbor agent of current intersection, $a_j$ is the neighbor action, and $m$ is the latent variable generated by the encoder.

\item
\textbf{Policy. } The policy takes local observations $o_{t}$ and latent variable as input, parameterised by $\theta$ and dependent on $\psi$, which is defined by
\begin{align}
    \pi^{\theta}\left(a \mid o_{t}, m\right),
\end{align}
where $m$ is derived by encoder.
The policy is conditioned on both observation and posterior over $m$, which is similar to the formulation of Bayesian RL \cite{zhou2019bayesian,poupart2006analytic,ghavamzadeh2015bayesian}.

\end{itemize}

\subsection{Latent Variable} \label{sec:policies_with_latent_variables}

\subsubsection{Design}

From the perspective of meta-RL, there exists a true variable to represent a certain task (i.e., an agent). However, we do not have access to this information. Alternatively, we aim to learn variable $m$ from trajectories of the current task.

Considering the uncertainty of the task, we model the task as a Gaussian distribution $\mathcal{N}(\mu,\sigma^2)$, where the mean $\mu$ and the standard deviation $\sigma$ are generated by the encoder. Specifically, a RNN is employed to track the characteristics of trajectory $\tau_{:t}$ and calculate  $\mu$ and $\sigma$ respectively. Then, $m$ is randomly sampled from $\mathcal{N}(\mu,\sigma^2)$, that is, $m\sim\mathcal{N}(\mu,\sigma^2)$.
Note $m\sim\mathcal{N}(\mu,\sigma^2)$ is underivable, we alternatively use the reparameterization technique \cite{kingma2013auto} for backpropagation:
\begin{align}
    m=\mu + \sigma m_0, 
\end{align}
where $m_0$ is randomly sampled from the standard normal distribution $m_0\sim {N}(\mu,I)$, and $I$ is the identity matrix.

\subsubsection{Analysis}

Besides making the meta-learning possible as shown in Eq.~\ref{eq:framework}, the latent variable brings two additional advantages:

\vspace{0.1cm}
\noindent \textbf{(1) Trade-off of Exploration and Exploitation: }
Latent variable is an additional input of the policy network except for the observation, which could be regarded as prior knowledge about the task. Previous research \cite{gupta2018meta} has shown that random Gaussian distribution could be regarded as the noise disturbance and provide randomness of the policy. Specifically, the regular policy network outputs action probability based on the current observation only, and the distribution of sampled action is certain once the current observation is given. In contrast, our method takes the latent variable $m$ as the additional input of 
 the policy network besides the observation, where $m$ is randomly sampled from the learnable Gaussian distribution $\mathcal{N}(\mu,\delta^2)$.
Hence, $m$ could bring  randomness to  the distribution of the sampled action in the learning procedure. Hence, it helps to explore the tasks. 

\begin{figure}[t]
 		\centering
 		\includegraphics[width=0.47\textwidth]{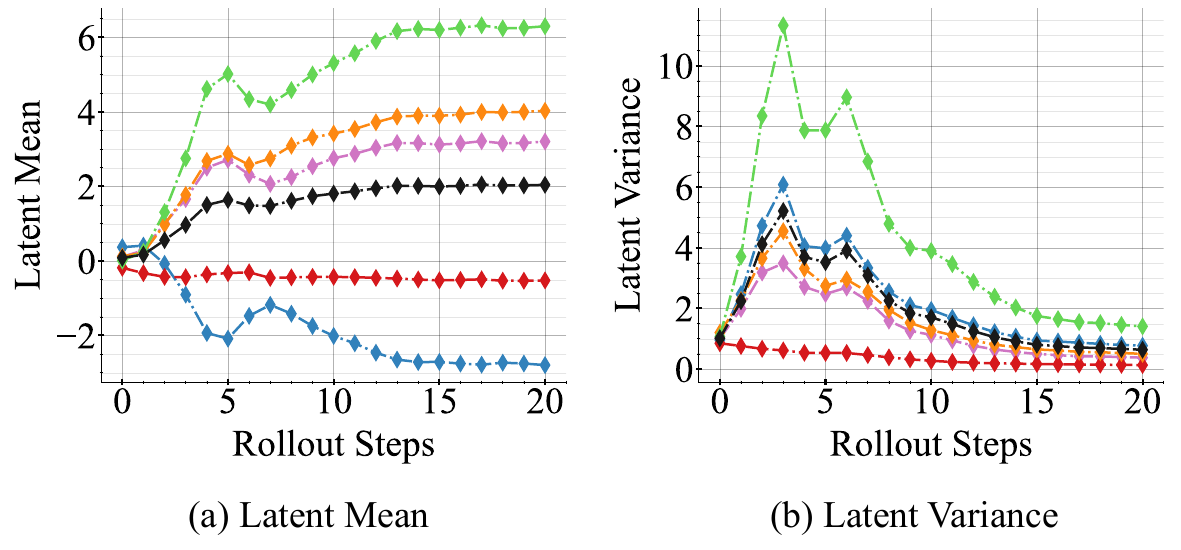} 
 		\vspace*{-0.2cm}
 		\caption{Visualisation of the latent space. (a) and (b) denote the mean and variance of the latent variable respectively. The latent variable has 5 dimensions, each line is one latent dimension, the black line is the average.}
 		\label{fig:latent_img}
 		\vspace*{-0.3cm}
\end{figure}

Since our Gaussian distribution changes dynamically in training rather than keeps as constant. It not only helps to explore but also helps to trade off the exploration and exploitation \cite{zintgraf2019varibad}. 
With the converging of the model, the randomness gradually decreases. Specifically, the mean $\mu$ tends to be stable and the variance $\delta^2$ tends to zero. 
Fig. \ref{fig:latent_img} illustrates the adaptation of latent variable. It has a dynamic guiding effect on policy: randomness is the largest at the beginning, the dispersive distribution of $m$ necessitates the agent to perform diverse actions to explore, and then randomness decreases as the posterior distribution $\mathcal{N}(\mu,\sigma^2)$ gradually converges. In this phase, the policy network encourages the agent to exploit the environment.
In addition, it also makes the final learned policy stable with very little randomness.
This can be analogous to the impact of $\epsilon$-greedy in DQN on trade-off of the exploration and exploitation.
The intuitive idea is: random noise (random latent variable) is equivalent to the fixed $\epsilon$ in $\epsilon$-greedy, and the learnable Gaussian distribution (learnable latent variable) is equivalent to the adjustable $\epsilon$ in $\epsilon$-greedy, which has a dynamic guiding effect on the trade-off of the exploration and exploitation.

\vspace{0.1cm}
\noindent \textbf{(2) Modelling Latent Coordination with Neighbors:} In our method, the neighbor information is available only in training, and the decoders are abandoned in execution. Only using individual observation as the input of policy may ignore the latent neighbors information. As shown in Eq.~\ref{eq:framework}, the observation transition is caused by not only the current agent but also its neighbors. In turn, the latent variable could reflect the latent neighbor's information.                               
\begin{figure}[t]
 		\centering
 		\includegraphics[width=0.45\textwidth]{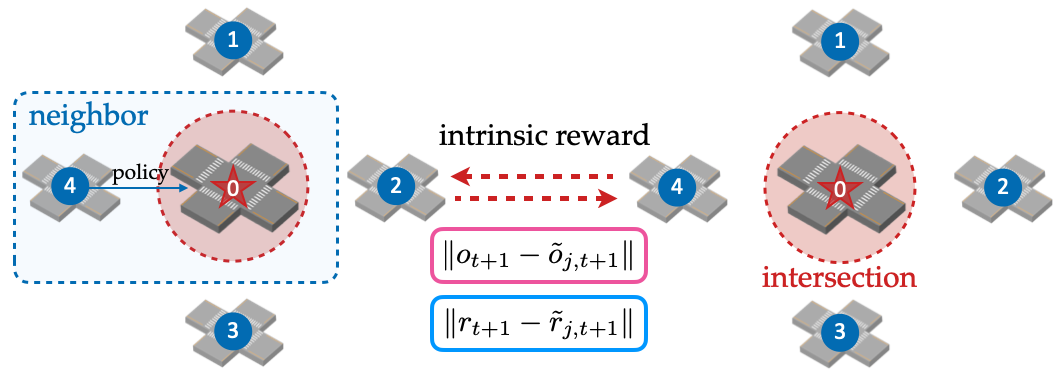} 
 		\caption{Illustration of intrinsic reward design.
 		}
 		\label{fig:i_r_d}
 		\vspace{-0.3cm}
\end{figure}

\subsection{Intrinsic Reward} \label{sec:intrinsic_reward}

\subsubsection{Design}

As stated in Eq.~\ref{eq:goal}, the non-stationary learning often causes the observation transition and received rewards unpredictable only conditioned on individual observation and action. Conversely, we hope the learned policy makes them be predicted stably.
To achieve this goal, we design a novel intrinsic reward based on VAE, as illustrated in Fig. \ref{fig:i_r_d}, which is the negative of the prediction bias with/without neighbor agents' policies:
\begin{align}
    r_{t}^{i n t}=-\sum_{j}\big(\left\|r_{t+1}-\tilde{r}_{j, t+1}\right\|+\left\|o_{t+1}-\tilde{o}_{j, t+1}\right\|\big),
    \label{eq:intrinsic_reward}
\end{align}
where  $r_{t+1}$ and $\tilde{r}_{j, t+1}$ are the predicted rewards with/without neighbor agents' policies respectively, $o_{t+1}$ and $\tilde{o}_{j, t+1}$ are predicted next observations with/without neighbor agent's policies respectively. They are generated by the corresponding decoders in Eq.~\ref{eq:decoder}.

\subsubsection{Analysis}
Here, will give an interpretation from the perspective of information theory why the intrinsic reward design can lead to a stable policy robust to neighbors’ policy, which could make the learned policy easy to transfer. 
From the perspective of the current agent, the $r_{t+1}$ is defined as
\begin{align}
    r_{t+1} &= \mathcal{R}\left(o_{t+1}, o_{t}, \mathbf{u}_{t}, a_{t}; m \right), 
\end{align}
for any neighbor agent $j$, the unobservable part $\mathbf{u}_{t}$ consists of two parts, the part containing $j$ and the part not containing $j$, that is $\mathbf{u}_{t} = \{o_{j, t}, \mathbf{u}_{t}^{-j}\}$, and $a_{j,t} \sim \pi_j$. Then add the neighbor agent's policy, we can obtain:
\begin{align}
    \tilde{r}_{j, t+1} = \mathcal{R}(o_{t+1}, o_{t}, \underbrace{o_{j, t}, \mathbf{u}_{t}^{-j}}_{\rm unobs.}, a_{t}, a_{j, t}; m).
\end{align}
Similar,  $o_{t+1}$ is defined as
\begin{align}
    o_{t+1} = \mathcal{T}\left(o_{t+1} \mid o_{t}, a_{t}, \mathbf{u}_{t} ; m\right),
\end{align}
we split the unobservable part $\mathbf{u}_{t+1}$ into the part containing $j$ and the part not containing $j$, then add the neighbor agent's policy, we have
\begin{align}
    \tilde{o}_{j, t+1} = \mathcal{T}(o_{t+1} \mid o_{t}, a_{t}, \underbrace{o_{j,t}, \mathbf{u}_{t}^{-j}}_{\rm unobs.}, a_{j,t} ; m).
\end{align}
According to Eq. \ref{eq:intrinsic_reward}, our intrinsic reward is:
\begin{align}
    \label{eq:inreward}
    r_{t}^{int} &=-\sum_{j}\big(\left\|r_{t+1}-\tilde{r}_{j, t+1}\right\|+\left\|o_{t+1}-\tilde{o}_{j, t+1}\right\|\big) \\ \nonumber
    &=-\sum_{j}\Big(\big\|\mathcal{R}\left(o_{t+1}, o_{t}, \mathbf{u}_{t}, a_{t} ; m \right) -  \\ \nonumber
    & \qquad \qquad \quad \mathcal{R}(o_{t+1}, o_{t}, o_{j, t}, \mathbf{u}_{t}^{-j}, a_{t}, a_{j, t} ; m) \big\|_{2}^{2}  \\ \nonumber
    & \qquad  + \big\|  \mathcal{T}\left(o_{t+1} \mid o_{t}, a_{t}, \mathbf{u}_{t} ; m \right) - \\ \nonumber
    &  \qquad  \qquad \quad \mathcal{T}(o_{t+1} \mid o_{t}, a_{t}, o_{j, t}, \mathbf{u}_{t}^{-j}, a_{j, t} ; m) \big\|_{2}^{2} \Big)\\ \nonumber
    & =-\sum_{j}\Big(\big\| \mathcal{R}\left(r_{t+1} \mid z_{t}^{2}\right)-\mathcal{R}\left(r_{t+1} \mid a_{j, t}, z_{t}^{2}\right)\big\|_{2}^{2} \ + \\ \nonumber
    & \qquad \qquad \ \  \big\| \mathcal{T}\left(o_{t+1} \mid z_{t}^{1}\right)-\mathcal{T}\left(o_{t+1} \mid a_{j, t}, z_{t}^{1}\right)\big\|_{2}^{2}\Big) 
\end{align}
where $z_{t}^{1}$ represents the conditioning variable 1 at timestep $t$, 
$z_{t}^{1}=\left\langle o_{t}, a_{t}, \mathbf{u}_{t}, m \right\rangle$, 
and $z_{t}^{2}$ represents the conditioning variable 2 at timestep $t$,
$z_{t}^{2}=\left\langle o_{t+1}, o_{t}, \mathbf{u}_{t}, a_{t}, m \right\rangle$.
The mutual information(MI) between the predicted state $o_{t+1}$ and another agent’s policy $a_{j}$ is:
\begin{align}
    \label{eq:item1}
    I & \left(  o_{t+1} ; a_{j,t} \mid z_{t}^{1}\right) \\
    &=\sum_{o_{t+1}, a_{j,t}} p\left(o_{t+1}, a_{j,t} \mid z_{t}^{1}\right) \log \frac{p\left(o_{t+1}, a_{j,t} \mid z_{t}^{1}\right)}{p\left(o_{t+1} \mid z_{t}^{1}\right) p\left(a_{j,t} \mid z_{t}^{1}\right)} \nonumber \\
    &=\sum_{ o_{t+1} ,a_{j,t}} p\left(o_{t+1} \mid a_{j,t}, z_{t}^{1}\right) \cdot p\left(a_{j,t} \mid z_{t}^{1}\right) \cdot \nonumber \\
    & \qquad \log \frac{p\left(o_{t+1} \mid a_{j,t}, z_{t}^{1}\right) \cdot p\left(a_{j,t} \mid z_{t}^{1}\right)}{p\left(o_{t+1} \mid z_{t}^{1}\right) p\left(a_{j,t} \mid z_{t}^{1}\right)} \nonumber \\
    &=\sum_{a_{j,t}} p\left(a_{j,t} \mid z_{t}^{1}\right) D_{\mathrm{KL}}\left[p\left(o_{t+1} \mid a_{j,t}, z_{t}^{1}\right) \| p\left(o_{t+1} \mid z_{t}^{1}\right)\right] \nonumber \\
    &=\sum_{a_{j,t}} p\left(a_{j,t} \mid z_{t}^{1}\right) D_{\mathrm{KL}}\left[T\left(o_{t+1} \mid a_{j,t}, z_{t}^{1}\right) \| T\left(o_{t+1} \mid z_{t}^{1}\right)\right]. \nonumber
\end{align}

The mutual information(MI) between the predicted reward $r_{t+1}$ and another agent’s policy $a_{j, t}$ is as follows:
\begin{align}
    \label{eq:item2}
    I & \left(r_{t+1} ; a_{j,t} \mid z_{t}^{2}\right)  \\
    &=\sum_{r_{t+1}, a_{j,t}} p\left(r_{t+1}, a_{j,t} \mid z_{t}^{2}\right) \log \frac{p\left(r_{t+1}, a_{j,t} \mid z_{t}^{2}\right)}{p\left(r_{t+1} \mid z_{t}^{2}\right) p\left(a_{j,t} \mid z_{t}^{2}\right)} \nonumber \\ 
    &=\sum_{r_{t+1} ,a_{j,t}} p\left(r_{t+1} \mid a_{j,t}, z_{t}^{2}\right) \cdot p\left(a_{j,t} \mid z_{t}^{2}\right)\nonumber \\
    &\qquad \log \frac{p\left(r_{t+1} \mid a_{j,t}, z_{t}^{2}\right) \cdot p\left(a_{j,t} \mid z_{t}^{2}\right)}{p\left(r_{t+1} \mid z_{t}^{2}\right) p\left(a_{j,t} \mid z_{t}^{2}\right)} \nonumber \\
    &=\sum_{a_{j,t}} p\left(a_{j,t} \mid z_{t}^{2}\right) D_{\mathrm{KL}}\left[p\left(r_{t+1} \mid a_{j,t}, z_{t}^{2}\right) \| p\left(r_{t+1} \mid z_{t}^{2}\right)\right] \nonumber \\
    &=\sum_{a_{j,t}} p\left(a_{j,t} \mid z_{t}^{2}\right) D_{\mathrm{KL}}\left[R\left(r_{t+1} \mid a_{j,t}, z_{t}^{2}\right) \| R\left(r_{t+1} \mid z_{t}^{2}\right)\right]. \nonumber
\end{align}
Combine the Eq. \ref{eq:item1} and Eq. \ref{eq:item2}, we have
\begin{align}
     & \qquad I\left(o_{t+1} ; a_{j,t} \mid z_{t}^{1}\right) + I\left(r_{t+1} ; a_{j,t} \mid z_{t}^{2}\right)  \\
    &=\sum_{a_{j,t}} p\left(a_{j,t} \mid z_{t}^{1}\right) D_{\mathrm{KL}}\left[T\left(o_{t+1} \mid a_{j,t}, z_{t}^{1}\right) \| T\left(o_{t+1} \mid z_{t}^{1}\right)\right] \nonumber \\
    & + \sum_{a_{j,t}} p\left(a_{j,t} \mid z_{t}^{2}\right) D_{\mathrm{KL}}\left[R\left(r_{t+1} \mid a_{j,t}, z_{t}^{2}\right) \| R\left(r_{t+1} \mid z_{t}^{2}\right)\right], \nonumber
\end{align}

By sampling $N$ independent trajectories $\tau_{N}$ from the environment, we perform a Monte-Carlo approximations of the MI: 
\begin{align}
    & \quad I\left(o_{t+1} ; a_{j,t} \mid z_{t}^{1}\right) + I\left(r_{t+1} ; a_{j,t} \mid z_{t}^{2}\right)  \\
    &=\mathbb{E_{\tau}} \bigg[ D_{\mathrm{KL}}\Big[T\left(o_{t+1} \mid a_{j,t}, z_{t}^{1}\right) \| T\left(o_{t+1} \mid z_{t}^{1}\right)\Big] \mid z_{t}^{1} \bigg] \ + \nonumber  \\
    & \qquad \mathbb{E_{\tau}}\bigg[ D_{\mathrm{KL}}\Big[R\left(r_{t+1} \mid a_{j,t}, z_{t}^{2}\right) \| R\left(r_{t+1} \mid z_{t}^{2}\right)\Big]\mid z_{t}^{2} \bigg], \nonumber \\
    & \approx \frac{1}{N} \sum_{N} \Big(\big\|T\left(o_{t+1} \mid a_{j,t}, z_{t}^{1}\right) - T\left(o_{t+1} \mid z_{t}^{1}\right)\big\| \ +  \nonumber \\
    &  \qquad \big\|R\left(r_{t+1} \mid a_{j,t}, z_{t}^{2}\right) - R\left(r_{t+1} \mid z_{t}^{2}\right)\big\|\Big). \nonumber
\end{align}

The last approximation is because the predictions in our implementation are deterministic rather than stochastic.

Thus, in expectation, the intrinsic reward is the negative of MI above. As each agent maximizes the long-term cumulative reward, which therefore minimizes MI. As a result, agents become independent. This can be an interpretation from the information-theoretical perspective. Note that the prediction results are only used to form intrinsic rewards, and our method tries to minimize them. That means our method mainly relies on the trend of change of predicted results, not the predicted value. Therefore, we expect our method is resilient to the decoders' modeling error accumulation.

\begin{algorithm}[t]
  \caption{MetaVIM: Meta-training Phase}
  \label{alg:MetaVIM_train}
\begin{algorithmic}
  \STATE {\bfseries Require:} A set of meta-training tasks $\{{\kappa}_i\}_{i \in \mathcal{N}}$ drawn from intersections $\mathcal{N}=1,2, \ldots, N$ in a multi-agent scenario, number of training episodes $E$
  \STATE Initialize policy replay buffers $\mathcal{B}^{\pi}$
  \STATE Initialize mVAE replay buffers $\mathcal{B}^{\mathrm{mVAE}}$
  \STATE Initialize encoder $\mathbf{e}^{\psi}$, decoders $\mathbf{p}^{\phi_{\mathrm{r}}}, \mathbf{p}^{\phi_{\o}}, \mathbf{p}^{\phi \tilde{\mathrm{r}}}, \mathbf{p}^{\phi_{\tilde{o}}}$
  \STATE Initialize policy $\pi^{\theta}$ 
  \WHILE{not done} 
  \FOR[Data collection]{episodes$=1,2, \ldots, E$}
  \STATE Compute latent variable $m$ from the current rollouts
  \STATE Add latent variable $m$ to $\mathcal{B}^{\pi}$

  \FOR[Training]{steps in training steps}
  \STATE Take action according to $\pi^{\theta}$
  \STATE Get neighbor actions $\left\{a_{j}\right\}_{j=1, \ldots, J}$
  \STATE Act to the environment and get environmental reward $r$
  \STATE Compute intrinsic reward $r^{int}$ using Eq. \eqref{eq:intrinsic_reward}
  \STATE Update latent variable $m$ by $\mathbf{e}^{\psi}$ 
  \STATE Add trajectories $\tau$ and latent variable $m$ to $\mathcal{B}^{\mathrm{mVAE}}$
  \STATE Add trajectories $\tau$ to $\mathcal{B}^{\pi}$
  \STATE Train encoder $\mathbf{e}^{\psi}$ and decoders $\mathbf{p}^{\phi_{\mathrm{r}}}, \mathbf{p}^{\phi_{\o}}, \mathbf{p}^{\phi \tilde{\mathrm{r}}}, \mathbf{p}^{\phi_{\tilde{o}}}$ by maximising ELBO in Eq. \eqref{eq:elbos}
  \STATE Train policy $\pi^{\theta}$ by maximizing reward in Eq. \eqref{eq:RL}
  \STATE clean up $\mathcal{B}^{\pi}$
  \ENDFOR
  \ENDFOR
  \ENDWHILE
\end{algorithmic}
\end{algorithm}

\begin{algorithm}[t]
  \caption{MetaVIM: Meta-testing Phase}
  \label{alg:MetaVIM_test}
\begin{algorithmic}
  \STATE {\bfseries Require:} A set of meta-testing tasks $\{{\kappa}_i\}_{i \in \mathcal{M}}$ drawn from intersections $\mathcal{M}=1,2, \ldots, M$ in a multi-agent scenario, number of testing episodes $E$
  \STATE Load policy model $\pi^{\theta}$
  \STATE Load encoder model $\mathbf{e}^{\psi}$ 
  \FOR[Rollouts]{episodes=1,2, ...,$E$}
  \STATE Compute latent variable $m$ from the current rollouts
  \FOR{step in testing steps}
  \STATE Take action according to $\pi^{\theta}$
  \STATE Act to the environment
  \STATE Compute latent variable $m$ by $\mathbf{e}^{\psi}$ 
  \ENDFOR
  \ENDFOR
  \FOR[Testing]{step in testing steps}
  \STATE Take action according to $\pi^{\theta}(a \mid o, m)$
  \ENDFOR
\end{algorithmic}
\end{algorithm}

\subsection{Learning}
The parameters of MetaVIM consist of a policy network $\pi^{\theta}\left(a_{t} \mid o_{t}, m\right)$ and a mVAE which includes a encoder $\mathbf{e}^{\psi}\left(m \mid \tau_{: t}\right)$ and four decoders $\mathbf{p}^{\phi_{\mathrm{r}}}\left(r_{t+1} \mid o_{t+1}, o_{t}, a_{t}, m\right)$, $\mathbf{p}^{\phi_{o}}\left(o_{t+1} \mid o_{t}, a_{t}, m\right)$, $\mathbf{p}^{\phi \tilde{\mathrm{r}}}\left(\tilde{r}_{t+1} \mid o_{t+1}, o_{t}, a_{t}, a_{j, t}, m\right)$ and $\mathbf{p}^{\phi_{\bar{o}}}\left(\tilde{o}_{t+1} \mid o_{t}, a_{t}, a_{j, t}, m\right)$, where $\theta, \psi, \phi_{\mathrm{r}}, \phi_{\mathrm{o}}, \phi_{\tilde{r}}$ and $\phi_{\tilde{o}}$ are the corresponding network weights. The learning mainly contains two parts: 

\vspace{0.1cm}
\noindent \textbf{(1) Maximizing the Cumulative Rewards:} 
For the policy network $\theta$, the learning objective is to maximize the cumulative reward as well as the intrinsic reward $r^{int}_t$ in Eq.~\ref{eq:intrinsic_reward}:
\begin{align}
    \max\limits_{\theta}J(\theta)=\mathbb{E}_{\substack{a_t \sim \pi^\theta(a_t|o_t,m) \\ m \sim \mathcal{N}(\mu, \sigma^2)}}\sum\limits_{t=0}^\mathcal{H}(r_t+\alpha r^{int}_t),
    \label{eq:RL}
\end{align}
where $\alpha$ is a positive weight. To achieve Eq.~\ref{eq:RL}, the Proximal Policy Optimization (PPO) \cite{schulman2017proximal} is employed. PPO is an on-policy actor-critic method which adds a soft constraint that can be optimized by a first-order optimizer, effectively using existing data to take step that leads to the biggest possible improvement on a policy, without stepping too far to accidentally cause performance collapse.

\vspace{0.1cm}
\noindent \textbf{(2) Training mVAE}: 
For current agent, the past trajectory up to time step $t$ is collected as
\begin{align}
    \tau_{ : t}=\left(o_{0}, a_{0}, r_{1}, o_{1}, a_{1}, r_{2}, \ldots, o_{t-1}, a_{t-1}, r_{t}, o_{t}\right),
\end{align}
where $a_{t} \sim \pi^{\theta}\left(a_{t} \mid o_{t}, \mathbf{e}^{\psi}\left(m \mid \tau_{: t}\right)\right)$. 
Based on $\tau_{: t}$, at any given time step $t$, our mVAE learning objective is thus to maximise
\begin{align}
    \mathbb{E}_{\tau_{:t} \sim \rho}\left[ \log \mathbf{p}^{\phi_r,\phi_{\tilde r},\phi_o,\phi_{\tilde o}}(\tau_{:t}) \right],
    \label{eq:mVAE_goal}
\end{align}
Since the original optimized objective of VAE can't be optimized directly, instead of optimising Eq. \ref{eq:mVAE_goal}, we use a variational evidence lower bound (ELBO) \cite{kingma2013auto} which is widely used to optimize the VAE-based network. Maximizing ELBO can approximately maximize the $\log \mathbf{p}^{\phi_r,\phi_{\tilde r},\phi_o,\phi_{\tilde o}}(\tau_{:t})$, the ELBO can be derived as follows:
\begin{align}
\label{eq:elbos_derivation}
    & \mathbb{E}_{\tau_{ :t} \sim \rho }\left[ \log \mathbf{p}^{\phi_r,\phi_{\tilde r},\phi_o,\phi_{\tilde o}}(\tau_{:t}) \right]  \\ 
    & 
    = \mathbb{E}_{\tau_{:t} \sim \rho }\left[ {\log \mathbf{p}^{\phi_r}(r_{t+1}) + \log \mathbf{p}^{\phi_{\tilde r}}(r_{t+1})  }\right. \nonumber \\
    & \qquad \left.{ \quad +  \log \mathbf{p}^{\phi_o}(o_{t+1}) + \log \mathbf{p}^{\phi_{\tilde o}}(o_{t+1}) }\right] \nonumber \\
    & 
    = \mathbb{E}_{\tau_{:t} \sim \rho }\left[ {\log \mathbb{E}_{\mathbf{e}_\psi(m |\tau_{:t})} [ { \frac{\mathbf{p}^{\phi_r}(r_{t+1}, m)}{\mathbf{e}_\psi(m |\tau_{:t})} }}  \right. \nonumber \\
    & \qquad \left.{ \quad +
    \frac{\mathbf{p}^{\phi_{\tilde r}}(r_{t+1}, m)}{\mathbf{e}_\psi(m |\tau_{:t})} +
    \frac{\mathbf{p}^{\phi_o}(o_{t+1}, m)}{\mathbf{e}_\psi(m |\tau_{:t})} + \frac{\mathbf{p}^{\phi_{\tilde o}}(o_{t+1}, m)}{\mathbf{e}_\psi(m |\tau_{:t})} ] }\right] \nonumber \\
    & 
    \ge \mathbb{E}_{\tau_{:t} \sim \rho , ~ \mathbf{e}_\psi(m |\tau_{:t})} \left[ {\log \frac{\mathbf{p}^{\phi_r}(r_{t+1}, m)}{\mathbf{e}_\psi(m |\tau_{:t})} } +  \log \frac{\mathbf{p}^{\phi_{\tilde r}}(r_{t+1}, m)}{\mathbf{e}_\psi(m |\tau_{:t})} \right. \nonumber \\
    & \qquad \left.{ \quad + \log \frac{\mathbf{p}^{\phi_o}(o_{t+1}, m)}{\mathbf{e}_\psi(m |\tau_{:t})} + \log \frac{\mathbf{p}^{\phi_{\tilde o}}(o_{t+1}, m)}{\mathbf{e}_\psi(m |\tau_{:t})} } \right] \nonumber \\
    & 
    = \mathbb{E}_{\tau_{:t} \sim \rho , ~ \mathbf{e}_\psi(m |\tau_{:t})} \left[ {\log \mathbf{p}^{\phi_r}(r_{t+1} | m)  +  \log \mathbf{p}^{\phi_{\tilde r}}(r_{t+1} | m) }\right. \nonumber \\
    &  \left.{ \quad + \log \mathbf{p}^{\phi_o}(o_{t+1} | m) + \log \mathbf{p}^{\phi_{\tilde o}}(o_{t+1} | m) - \log \mathbf{e}_\psi(m |\tau_{:t}) }\right] \nonumber \\
    & 
    = \mathbb{E}_{ \mathbf{e}_\psi(m |\tau_{:t})}\sum\limits_{t'=0}^{t-1}( \log \mathbf{p}^{\phi_r}(r_{t+1}) +\log \mathbf{p}^{\phi_{\tilde r}}(r_{t+1}) \nonumber \\
    & +\log \mathbf{p}^{\phi_o}(o_{t+1})+\log \mathbf{p}^{\phi_{\tilde o}}(o_{t+1}))
            - KL(\mathbf{e}_\psi(m |\tau_{:t}) || q(m )).  \nonumber \\
    &
    = ELBO(\psi,\phi_r,\phi_{\tilde r},\phi_o,\phi_{\tilde o}|\tau_{:t}). \nonumber
\end{align}

\begin{figure}[t]
 		\centering
 		\includegraphics[width=0.49\textwidth]{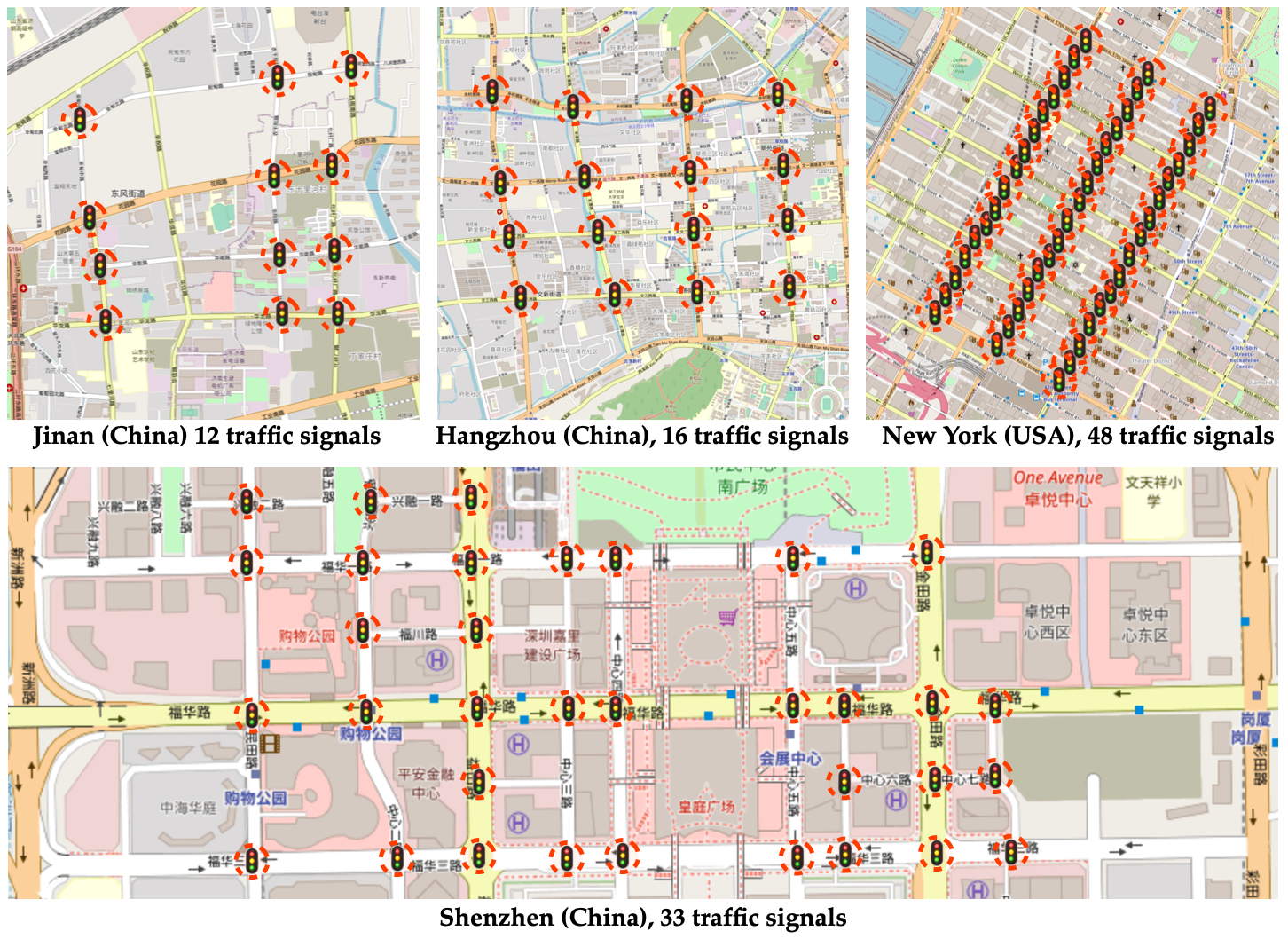} 
 		\vspace*{-0.4cm}
 		\caption{The illustration of the road networks. The first row shows the road networks of Jinan (China), Hangzhou (China) and New York (USA), containing 12, 16 and 48 traffic signals respectively, and the second row shows the road network of Shenzhen containing 33 traffic signals. }
 		\label{fig:road_network}
 		\vspace*{-0.1cm}
\end{figure}

Then the mVAE learning objective is thus to maximise:
\begin{align}
    &ELBO(\psi,\phi_r,\phi_{\tilde r},\phi_o,\phi_{\tilde o}|\tau_{:t}) \nonumber \\
    &=\mathbb{E}_{ \mathbf{e}^\psi(m |\tau_{:t})}\sum\limits_{t=0}^{t-1}( \log \mathbf{p}^{\phi_r}(r^\tau_{t+1}) +\log \mathbf{p}^{\phi_{\tilde r}}(r^\tau_{t+1}) \nonumber \\
    &\qquad \quad +\log \mathbf{p}^{\phi_o}(o^\tau_{t+1})+\log \mathbf{p}^{\phi_{\tilde o}}(o^\tau_{t+1}))\nonumber \\
    &\qquad \quad - KL(\mathbf{e}^\psi(m |\tau_{:t}) || q(m)).
    \label{eq:elbos}
\end{align}

The first 4 terms are often referred to as the reconstruction loss, and the term $KL(\mathbf{e}^\psi(m|\tau_{:t}) || q(m))$ is the KL-divergence between the variational posterior and the prior distribution $q(m)$,  which is set to standard normal distribution $\mathcal{N}(0,\mathcal{I})$ initially, where $\mathcal{I}$ means the identity matrix. 
Suppose $\rho_i$ is the trajectory distribution induced by our policy and the transition function $\mathcal{T}(\cdot)$, then the learning objective of the mVAE is to maximize the ELBO over $\rho$:
\begin{align}
    \max\limits_{\psi,\phi_r,\phi_{\tilde r},\phi_o,\phi_{\tilde o}}\mathbb{E}_{\tau_{:t} \sim \rho }ELBO(\psi,\phi_r,\phi_{\tilde r},\phi_o,\phi_{\tilde o}|\tau_{:t}).
    \label{eq:elbos1}
\end{align}

The overall training algorithm is provided in Alg. \ref{alg:MetaVIM_train}. In our experiments, we train the policy and the mVAE using different replay buffers: $\mathcal{B}^{\pi}$ only collects recent trajectories since we use on-policy RL algorithms, and for the $\mathcal{B}^{\text{mVAE}}$ we maintain a larger buffer of observed trajectories. At meta-test time, we roll out the policy to evaluate the performance, meta-testing procedure is provided in Alg. \ref{alg:MetaVIM_test}. The decoders are not used at test time, and no gradient adaptation is done: $\theta$ and $\psi$ are shared across different tasks and the policy has learned to act approximately optimal during meta-training. See Appendix \ref{sec:implementation_details} for implementation details.



\section{Experiments}\label{sec:experiments}

We conduct the experiments on CityFlow \cite{zhang2019cityflow}, an city-level open-source simulation platform for traffic signal control. The simulator is used as the environment to provide state for traffic signal control, the agents execute actions by changing the phase of traffic lights, and the simulator returns feedback. Specifically, we conduct additional experiments on another platform SUMO\footnote{\url{http://sumo.dlr.de/index.html}} and under different lane configurations to demonstrate the robustness of the method in Appendix \ref{sec:experiments_on_sumo} and Appendix \ref{sec:experiments_on_different_lane_configurations} respectively. 

\subsection{Datasets}

\subsubsection{Road Networks}
\label{sec:road_networks}

\begin{figure}[t]
 		\centering
 		\includegraphics[width=0.4\textwidth]{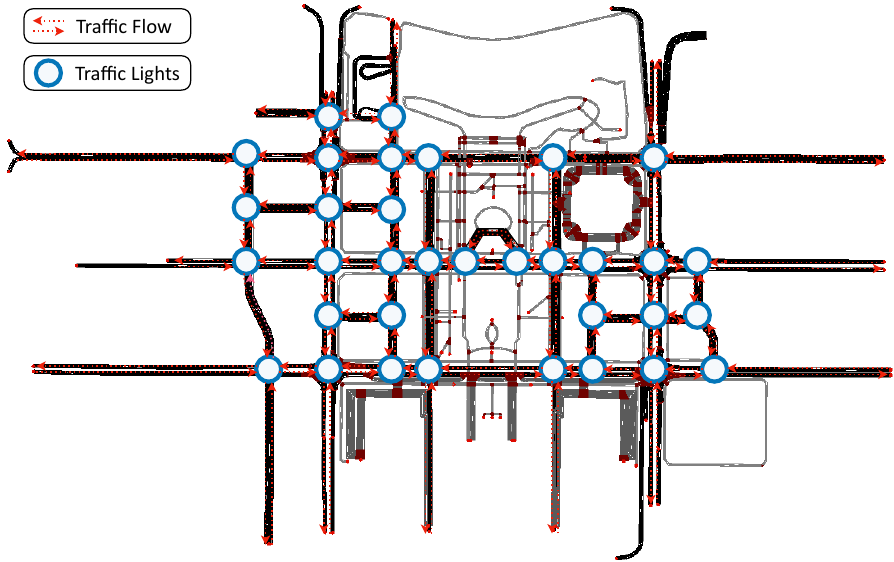} 
 		\vspace*{-0.25cm}
 		\caption{Overview of the Shenzhen (China) road network. There are total 33 traffic signal intersections across the area.}
 		\label{fig:fuhua_roadnet}
 		\vspace*{-0.2cm}
\end{figure}

The evaluation scenarios come from four real road network maps of different scales, including \textbf{Hangzhou} (China), \textbf{Jinan} (China), \textbf{New York} (USA) and \textbf{Shenzhen} (China), illustrated in Fig. \ref{fig:road_network}. The road networks and data of Hangzhou, Jinan and New York are from the public datasets\footnote{\url{https://traffic-signal-control.github.io/}}. The road network map of Shenzhen is made by ourselves which is derived from OpenStreetMap\footnote{\url{https://github.com/zhuliwen/RoadnetSZ}}. The road networks of Jinan and Hangzhou contain 12 and 16 intersections in $4 \times 3$ and $4 \times 4$ grids, respectively. The road network of New York includes 48 intersections in $16 \times 3$ grid. The road network of Shenzhen contains 33 intersections, which is not grid compared to other three maps, illustrated in Fig. \ref{fig:fuhua_roadnet}. In addition, an additional experiment are conducted on a much larger road network  to validate the scalability, more details are listed in Appendix \ref{sec:scalability_validation}. 

\label{sec:Scalability Validation}

\begin{figure}[h]
 		\centering
 		\includegraphics[width=0.48\textwidth]{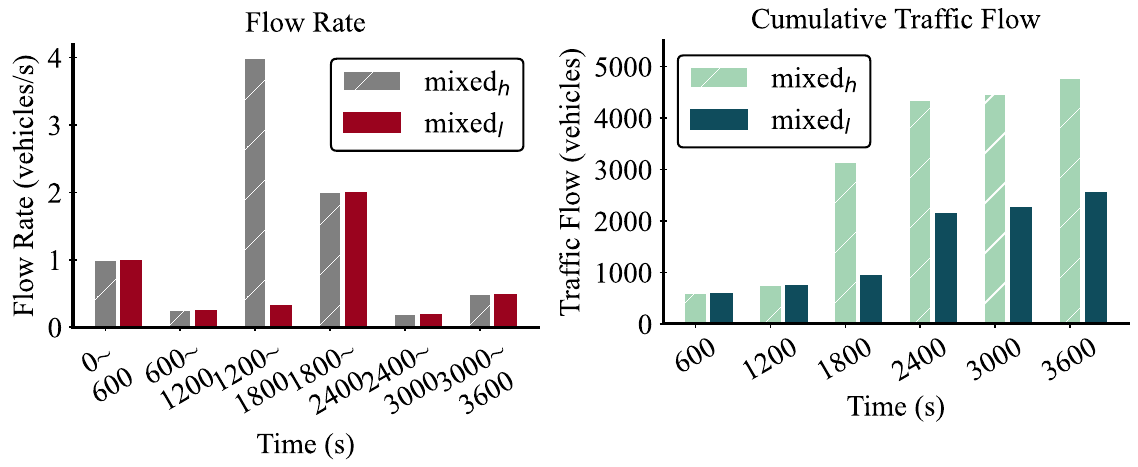}
 		\caption{Illustration of mixed flow. The left figure shows the arrival rate of vehicles, and the right figure shows the cumulative number of vehicles arriving in one hour.}
 		\label{fig:mixed_flow}
 		\vspace{-0.4cm}
\end{figure}

\subsubsection{Flow configurations}
\label{sec:flow_configurations}

We run the experiments under three traffic flow configurations: real traffic flow, mixed low traffic flow and mixed high traffic flow. The real traffic flow is real-world hourly statistical data with slight variance in vehicle arrival rates, as shown in Tab. \ref{tab:data_statistics_1}. Since the real-world strategies tend to break down during bottleneck period (peak hour), to better evaluate the performances of traffic light control methods in the flat-peak-flat scenario, we use two synthetic datasets: mixed high and mixed low traffic flow, which have a more dramatic variance in vehicle arrival rates, as shown in Tab. \ref{tab:data_statistics_3}. A detailed description of traffic flow configurations is:

\begin{itemize}[leftmargin=15pt]
\item
\textbf{\textit{Real}.} The traffic flows of \textbf{Hangzhou} (China), \textbf{Jinan} (China) and \textbf{New York} (USA) are from the public datasets\footnote{https://traffic-signal-control.github.io/}, which are processed from multiple sources. The traffic flow of \textbf{Shenzhen} (China) is made by ourselves generated based on the traffic trajectories collected from 80 red-light cameras and 16 monitoring cameras in a hour. The data statistics are listed in Tab. \ref{tab:data_statistics_1} and Tab. \ref{tab:data_statistics_2}.

\item
\textbf{{Mixed$_{l}$}.} The {mixed$_{l}$} is a mixed low traffic flow with a total flow of 2550 in one hour, to simulate a light peak. The arrival rate changes every 10 minutes, which is used to simulate the uneven traffic flow distribution in the real world, the details of the vehicle arrival rate and cumulative traffic flow are shown in Fig. \ref{fig:mixed_flow}.

\item
\textbf{{Mixed$_{h}$}.} The {mixed$_{h}$} is a mixed high traffic flow with a total flow of 4770 in one hour, in order to simulate a heavy peak. The difference from the mixed$_{l}$ setting is that the arrival rate of vehicles during 1200-1800s increased from 0.33 vehicles/s to 4.0 vehicles/s. The data statistics are listed in Tab. \ref{tab:data_statistics_3}. 
\end{itemize}

\renewcommand\tabcolsep{6.0pt}  
\begin{table}[h] 
    \footnotesize 
    \centering
    \setlength{\abovecaptionskip}{3pt}
    \caption{Arrival rate of real-world traffic dataset}
    \label{tab:data_statistics_1}
    \begin{tabular}{l l rrrr}
    \toprule
    \multirow{3}{*}{\textbf{Dataset}}
    & \multirow{3}{*}{\textbf{\# Intersections}}
    & \multicolumn{4}{c}{\textbf{Arrival rate (vehicles/300s)}} \\
    \cmidrule(lr){3-6}  
    & & \textbf{Mean} & \textbf{Std} & \textbf{Max} & \textbf{Min} \\
    \midrule
    Hangzhou & 16 (4 $\times$ 4) & 248.58 & 42.25 & 333 & 212 \\
    Jinan & 12 (4$\times$3) & 524.58 & 102.91 & 672 & 256 \\
    NewYork & 48 (16$\times$3) & 235.33 & 5.84 & 244 & 224 \\
    Shenzhen & 33 (Non-grid) & 147.92 & 79.35 & 255 & 22 \\
    \bottomrule
    \end{tabular}
    \vspace{-0.2cm}
\end{table}

\renewcommand\tabcolsep{7pt}  
\begin{table}[h] 
    \footnotesize 
    \centering
    \setlength{\abovecaptionskip}{3pt}
    \caption{Data statistics of synthetic traffic dataset}
    \label{tab:data_statistics_3}
    \begin{tabular}{l c rrrr}
    \toprule
    \multirow{3}{*}{\textbf{Dataset}}
    & \multirow{3}{*}{\textbf{Total vehicles}}
    & \multicolumn{4}{c}{\textbf{Arrival rate (vehicles/300s)}} \\
    \cmidrule(lr){3-6}  
    & & \textbf{Mean} & \textbf{Std} & \textbf{Max} & \textbf{Min} \\
    \midrule
    {mixed$_{l}$} & 2570 & 214.17 & 198.41 & 600 & 60 \\
    {mixed$_{h}$} & 4770 & 397.50 & 420.75 & 1200 & 60 \\
    \bottomrule
    \end{tabular}
\end{table}

\renewcommand\tabcolsep{6.5pt}  
\begin{table}[h] 
    \footnotesize 
    \centering
    \setlength{\abovecaptionskip}{3pt}
    \caption{Route statistics of real-world traffic dataset}
    \label{tab:data_statistics_2}
    \begin{tabular}{l l rrrr}
    \toprule
    \multirow{3}{*}{\textbf{Dataset}}
    & \multirow{3}{*}{\textbf{\# Intersections}}
    & \multicolumn{4}{c}{\textbf{Route (lanes/route)}} \\
    \cmidrule(lr){3-6}  
    & & \textbf{Mean} & \textbf{Std} & \textbf{Max} & \textbf{Min} \\
    \midrule
    Hangzhou & 16 (4 $\times$ 4) & 4.65	& 2.15 & 16 & 2 \\
    Jinan & 12 (4$\times$3) & 4.37 & 1.88 & 17 & 2 \\
    NewYork & 48 (16$\times$3) & 10.00 & 4.63 & 21 & 3 \\
    Shenzhen & 33 (Non-grid) & 7.57 & 3.91 & 41 & 1 \\
    \bottomrule
    \end{tabular}
    \vspace{-0.4cm}
\end{table}

\subsection{Evaluation Criteria}
\label{sec:evaluation_criteria}

Following existing studies \cite{wei2019colight,wei2019presslight,xiong2019learning,chen2020toward,zang2020metalight}, we use the \textbf{average travel time} to evaluate the performance of different methods for traffic signal control. The average travel time indicates the overall traffic situation in an area over a period of time. For a detailed definition of average travel time, see Section \ref{sec:preliminary}. Since the number of vehicles and the origin-destination (OD) positions are fixed, better traffic signal control strategies result in less average travel time.

\subsection{Testing Mode}
\label{sec:testing_mode}
The method is evaluated in two modes: \textbf{(1) Common Testing Mode}: the model trained on one scenario with one traffic flow configuration is tested on the same scenario with the same configuration. It is used to validate the ability of the RL algorithm to find the optimal policy. 
\noindent\textbf{(2) Meta-Test Mode}: we train the model in the Hangzhou road network and transfer the model to the other three networks directly. It is used to validate the generality of the model.

\subsection{Baselines}
\label{sec:baselines}
Our method is compared with the following two categories of methods: conventional transportation methods and RL methods\footnote{Some existing RL based methods, such as AttendLight \cite{oroojlooy2020attendlight} and SD-MaCAR \cite{guo2021urban}, evaluate their methodS under different experimental settings (e.g., road network or traffic flow), and the source codes are not available yet. Therefore, they are not compared.}. Note that for a fair comparison all the RL methods are learned without any pre-trained parameters and the methods are evaluated under the same settings. The results are obtained by running the source codes\footnote{\url{https://github.com/traffic-signal-control/RL_signals}}. All the baselines are run with three random seeds, and the mean is taken as the final result. The action interval is five seconds for each method, and the horizon is 3600 seconds for each episode. Specifically, the compared methods contain:

\subsubsection{Conventional methods}
\label{sec:conventional_methods}

\begin{itemize}[leftmargin=15pt]
\item
\textbf{Random} selects available phase randomly. 

\item
\textbf{MaxPressure \cite{varaiya2013max}} is a leading conventional method, which greedily chooses the phase with the maximum pressure. The pressure is defined as the difference of vehicle density between the incoming lane and the outgoing lane.

\item
\textbf{Fixedtime \cite{koonce2008traffic}} with random offset \cite{roess2004traffic} executes each phase in a phase loop with a pre-defined span of phase duration, which is widely used for steady traffic.

\item
\textbf{FixedtimeOffset \cite{koonce2008traffic}} where multiple intersections use the same synchronized fix-time plan. The offset is the time interval between the start time of the green light among intersections.

\item
\textbf{SlidingFormula \cite{roess2004traffic}} is designed based on the expert experience, which dynamically divides the duration of each phase according to the number of vehicle arriving.

\item
\textbf{SOTL \cite{cools2013self}} specifies a pre-defined threshold for the number of waiting vehicles on approaching lanes. Once the waiting vehicles exceeds the threshold, it will switch to the next phase.
\end{itemize}

\subsubsection{RL-based methods}
\label{sec:rl_based_methods}

\begin{itemize}[leftmargin=15pt]
\item
\textbf{Individual RL \cite{wei2018intellilight}} where each intersection is controlled by one agent. The replay buffer and network parameters are not shared, and the model update is independent. There is no information transfer between agents. 

\item
\textbf{MetaLight \cite{zang2020metalight}} is a value-based meta RL method via parameter initialization based on MAML \cite{finn2017model}. MetaLight is originally a single-agent approach for meta-learning on multiple separate tasks. Here we extend it to a multi-agent scenario without considering neighbor information.

\item
\textbf{PressLight \cite{wei2019presslight}} combines the traditional traffic method MaxPressure \cite{varaiya2013max} with RL  together, and optimizes the pressure of each intersection.

\item
\textbf{CoLight \cite{wei2019colight}} uses graph convolution and attention mechanism to model the neighbor information, and then further uses this neighbor information to optimize the queue length.

\item
\textbf{GeneraLight \cite{zhang2020generalight}} is a meta RL method which uses generative adversarial network to generate diverse traffic flows and
uses them to build training environments.
\end{itemize}

\subsection{Performance Comparison}
\label{sec:performance_comparion}

\renewcommand\arraystretch{1.05}
\renewcommand\tabcolsep{3.5pt}  

    \begin{table*}[t!]
    \footnotesize 
    \centering
    \setlength{\abovecaptionskip}{3pt}
    \caption{Performance on the Common Testing Mode. The smaller average travel times (s/veh) mean the better performances.}
    \label{tab:performance_1}
    \begin{tabular}{l || rrr  || rrr || rrr || rrr || r}
    \toprule
    \multirow{3}{*}{\textbf{Model}}
    & \multicolumn{3}{c||}{\textbf{Hangzhou}}
    & \multicolumn{3}{c||}{\textbf{Jinan}} 
    & \multicolumn{3}{c||}{\textbf{Newyork}} 
    & \multicolumn{3}{c||}{\textbf{Shenzhen}} 
    & \multirow{3}{*}{\textbf{Mean}} \\
    
    \cmidrule(lr){2-4} \cmidrule(lr){5-7} \cmidrule(lr){8-10} \cmidrule(lr){11-13}
    & \textbf{real} & \textbf{mixed$_{l}$} & \textbf{mixed$_{h}$} & \textbf{real} & \textbf{mixed$_{l}$}  &  \textbf{mixed$_{h}$} & \textbf{real} & \textbf{mixed$_{l}$}  &  \textbf{mixed$_{h}$} & \textbf{real} & \textbf{mixed$_{l}$}  &  \textbf{mixed$_{h}$} & \\
    \midrule
    Random & 727 & 1721 & 1794 & 836 & 1547 & 1733 & 1858 & 1865 & 2105 & 728 & 1775 & 1965 & 1554\\
    MaxPressure & 416 & 2449 & 2320 & \bf 355 & 839 & 1218 & 380 & 488 & 1481 & \bf 389 & 753 & 1387 & 1039\\
    Fixedtime & 718 & 1756 & 1787 & 814 & 1532 & 1739 & 1849 & 1865 & 2086 & 786 & 1705 & 1845 & 1540\\
    FixedtimeOffset & 736 & 1755 & 1725 & 854 & 1553 & 1720 & 1919 & 1901 & 2141 & 798 & 1886 & 2065 & 1588\\
    SlidingFormula & 441 & 1102 & 1241 & 576 & 759 & 1251 & 1096 & 986 & 1656 & 452 & 876 & 1347 & 982\\
    SOTL & 1209 & 2051 & 2062 & 1453 & 1779 & 1991 & 1890 & 1923 & 2140 & 1376 & 1902 & 2098 & 1823\\
    \midrule
    Individual RL & 743 & 1704 & 1819 & 843 & 1552 & 1745 & 1867 & 1869 & 2100 & 769 & 1753 & 1845 & 1551\\
    MetaLight & 480 & 1465 & 1576 & 784 & 984 & 1854 & 261 & 482 & 2145 & 694 & 954 & 2083 & 1147\\
    PressLight & 529 & 1538 & 1754 & 809 & 1173 & 1930 & 302 & 437 & 1846 & 639 & 834 & 1832 & 1135\\
    CoLight & 297 & 960 & 1077 & 511 & 733 & 1217 & 159 & 305 & 1457 & 438 & 657 & 1367 & 767\\
    GeneraLight & 335 & \bf 798 & 1575 & 586 & 1257 & 1616 & 1209 & 1257 & 1686 & 792 & 916 & 1574 & 1133 \\
    \midrule 
    Base & 526 & 1501 & 1674 & 793 & 1093 & 1904 & 298 & 432 & 1793 & 593 & 813 & 1732 & 1096 \\
    Base+$m$ & 302 & 923 & 1643 & 593 & 837 & 1532 & 152 & 298 & 1834 & 426 & 676 & 1630 & 906 \\
    Base+$m$+tran\_RS & 348 & 1134 & 1289 & 694 & 863 & 1406 & 189 & 368 & 1783 & 472 & 694 & 1416 & 888 \\
    Base+$m$+rew\_RS & 338 & 1049 & 1274 & 683 & 849 & 1375 & 173 & 327 & 1738 & 453 & 683 & 1372 & 860 \\
    MetaVIM &\bf 284 & 893 &\bf 986 & 492 &\bf 694 &\bf 1189 &\bf 149 &\bf 288 &\bf 1387 & 408 &\bf 622 &\bf 1272 & \bf724\\\bottomrule
    \end{tabular}
\end{table*}

\renewcommand\arraystretch{1.1}
\renewcommand\tabcolsep{5.0pt}  
    \begin{table*}[t!]
    \footnotesize 
    \centering
    \setlength{\abovecaptionskip}{3pt}
    \caption{Comparisons on the Meta-Test Mode. The \textit{original} means the model is trained on the current testing scenario, and the \textit{transfer} stands for the model is trained on the road map of Hangzhou. }
    \label{tab:performance_2}
    \begin{tabular}{l || rrr || rrr || rrr || c}
    \toprule
    \multirow{3}{*}{\textbf{Model}}
    & \multicolumn{3}{c||}{\textbf{Jinan}} 
    & \multicolumn{3}{c||}{\textbf{Newyork}} 
    & \multicolumn{3}{c||}{\textbf{Shenzhen}} 
    & \multirow{3}{*}{\textbf{Decline ratio}} \\
     \cmidrule(lr){2-4} \cmidrule(lr){5-7} \cmidrule(lr){8-10} 
    & \textbf{real} & \textbf{mixed$_{l}$} & \textbf{mixed$_{h}$} & \textbf{real} &  \textbf{mixed$_{l}$}  & \textbf{mixed$_{h}$} & \textbf{real} &  \textbf{mixed$_{l}$}  & \textbf{mixed$_{h}$} & \\
    \midrule
    Individual RL (origin) & 843 & 1552 & 1745 & 1867 & 1869 & 2100 & 769 & 1753 & 1845 & | \\
    \gr
    Individual RL (transfer) & 1198 & 2198 & 2493 & 2578 & 2330 & 2837 & 1046 & 2487 & 2513  & 38\% \\
    MetaLight (origin) & 784 & 984 & 1854 & 261 & 482 & 2145 & 694 & 954 & 2083 & | \\
    \gr
    MetaLight (transfer) & 983 & 982 & 2287 & 316 & 593 & 2487 & 865 & 1139 & 2593 & 19\% \\
    PressLight (origin) & 809 & 1173 & 1930 & 302 & 437 & 1846 & 639 & 834 & 1832 & | \\
    \gr
    PressLight (transfer) & 1119 & 1703 & 2693 & 429 & 569 & 2376 & 906 & 1287 & 2673 & 41\% \\
    CoLight (origin) & 511 & 733 & 1217 & 159 & 305 & 1457 & 438 & 657 & 1367 & | \\ 
    \gr
    CoLight (transfer) & | & | & | & | & | & | & | & | & | & | \\
    GeneraLight (origin) & 586 & 1257 & 1616 & 1209 & 1257 & 1686 & 792 & 916 & 1574 & | \\
    \gr
    GeneraLight (transfer) & 686 & 1571 & 2101 & 1330 & 1320 & 2057 & 935 & 1026 & 1747 & 17\% \\
    \midrule 
    MetaVIM (origin) &\bf 492 &\bf 694 &\bf 1189 &\bf 149 &\bf 288 &\bf 1387 &\bf 408  &\bf 622 &\bf 1272 &\bf | \\
    \gr
    MetaVIM (transfer) &\bf 513 &\bf 729 &\bf 1362 &\bf 153 &\bf 341 &\bf 1477 &\bf 443 &\bf 682 &\bf 1401 &\bf 9\% \\\bottomrule
    \end{tabular}
\end{table*}

\subsubsection{Evaluation on Common Testing}
\label{sec:evaluation_on_common_testing}

Tab.~\ref{tab:performance_1} lists the comparative results on the common testing mode, and it is evident that:

1) In general, RL methods perform better than conventional methods, and it indicates the advantage of the RL. The reason is that the conventional methods often rely on prior knowledge which may fails in some cases. A typical case is MaxPressure. It shows good performances on several cases including Hangzhou with the \textit{real} configuration, Jinan with the \textit{real,mixed$_l$} configurations,  NewYork with the \textit{real,mixed$_l$} configurations, and  Shenzhen with the \textit{real,mixed$_l$} configurations. However, it dramatically drops in other scenarios. That is, once the scenarios meet the assumption well, the method performs well and vice versa. 
Moreover, MetaVIM is comprehensively superior to other methods clearly in all scenarios and configurations, which demonstrates the effectiveness of the method.

2) MetaVIM shows good generalization for different scenarios and configurations. MetaVIM performs the second best in Hangzhou with the \textit{mixed$_l$} configuration, Jinan with the \textit{real} configuration and Shenzhen with the \textit{mixed$_l$} configuration, and performs best in other scenarios. Overall, MetaVIM has the best mean performance.
Except MaxPressure analysed above, GeneraLight achieves the best in Hangzhou with the \textit{mixed$_l$} configuration, while performs poorly in other scenarios. The reason is that GeneraLight  trains several models on  diverse generated traffic flows, and select the model in testing by matching the flow. Hence, it limits the generalization once the testing flow differs largely from the training flows.

3) MetaVIM outperforms Individual RL, MetaLight and PrssLight with 827, 423 and 411, respectively. The main reason is that they learn the traffic signal's policy only using its own observation and ignore the influence of the neighbors, while  MetaVIM considers the neighbors as the unobserved part of the current signal to help learning.  In addition, Individual RL performs relatively worst in all scenarios because it employs independent replay buffer and neural network parameters among agents. In traffic signal control task, different signals vary but also share similarities since they follow the same traffic rules and have similar optimization goals. Hence, sharing the replay buffer and neural network is helpful. 

4) The neighbors' information is modeled in CoLight and it performs well.It indicates modeling neighbors is critical for the coordination. The results of MetaVIM is superior to CoLight on each scenario and configuration, resulting mean 43 improvement. Compared to Colight, MetaVIM proposes an intrinsic reward to help the policies learning stable, and use latent variable to better trade off the exploration and exploitation. 
In addition, Colight needs the  agents' communications in testing, which is unnecessary in MetaVIM. It makes MetaVIM easy to deploy.

\subsubsection{Evaluation on Meta-Test}
\label{sec:evaluation_on_meta_test}

The comparative results evaluated on the meta-test mode are shown in Tab.~\ref{tab:performance_2}. The ``original'' means the model is trained on the current testing scenario, and the ``transfer'' stands for the model is trained on the road map of Hangzhou. From the results, we can obtain follow findings:

1) Colight needs full state information in both training and testing, hence it cannot be used for a new scenario which contains different number intersections compared with the training scenario. That is, the heterogeneous scenarios will cause heterogeneous inputs of the policy network, which makes the network fail to work. Hence, its results are unavailable in this setting. In contrast, the input of decentralized policy is the current agent's observation, and it is easy to adapt to a new scenario. Therefore, it is necessary to develop the decentralized policy for cross-scenario generalization.

2) The performances of Individual RL and PressLight drop 38\% and 41\% when the model is transferred. It shows that the models learned by the regular RL algorithms indeed rely on the training scenario. MetaLight is more robust to various scenarios than Individual RL and PressLight, and it indicates the advantage of the meta-learning framework. The meta-learning framework could help to learn task-shared model. Overall, MetaVIM achieves the state-of-the-art performances and only drops 9\% when transferring the model. The main reason is that: the  task-specific information is modeled by the latent variable in our method, and the learned policy function could be adaptive to diverse latent variables. That is, given a novel or unseen task, the task-specific information would be represented as latent variable rather than acting  as distractors. Hence, the latent variable helps to learn the across-task shared policy function better.

\subsubsection{Ablations}


To better validate the contribution of each component, four variants of MetaVIM are evaluated on the common testing mode in the Shenzhen road network, including:
\begin{itemize}[leftmargin=15pt]
\item
\textbf{Base} only keeps the policy network and removes the mVAE and latent variable $m$.

\item
\textbf{Base+$m$} where the encoder and $m$ are introduced. Base+$m$ contains policy network and encoder, which keeps latent variable but discards intrinsic rewards.

\item
\textbf{Base+$m$+tran\_RS} contains policy network, encoder and transition decoders but discards reward decoders. Transition decoders $\mathbf{p}^{\phi_{0}}\left(o_{t+1}\right)$ and $\mathbf{p}^{\phi_{\tilde{o}}}\left(\tilde{o}_{t+1}\right)$ are used and only $\left\|o_{t+1}-\tilde{o}_{j, t+1}\right\|$ is remained in Eq.~\ref{eq:inreward} to form the intrinsic reward.

\item
\textbf{Base+$m$+rew\_RS} contains policy network, encoder and reward decoders but discards transition decoders. Reward encoders $\mathbf{p}^{\phi_{\mathrm{r}}}\left(r_{t+1}\right)$ and $\mathbf{p}^{\phi \tilde{\mathrm{r}}}\left(\tilde{r}_{t+1}\right)$ are used and only $\left\|r_{t+1}-\tilde{r}_{j, t+1}\right\|$ is remained in Eq.~\ref{eq:inreward} to form the intrinsic reward.
\end{itemize}
Overall, MetaVIM contains the whole modules: policy network, encoder and decoders.

The qualitative evaluation results are listed in Tab. \ref{tab:performance_1} and the learning curves are shown in Appendix~\ref{sec:learning_curves}. 
We can obtain the following findings: 1) Among these 5 models, the performance of \textit{Baseline} is the worst. The reason is that it is hard to learn the effective decentralized policy independently in the multi-agent traffic signal control task, where one agent's reward and transition are affected by its neighbors. 2) Compared with the baseline, the improvement of  \textit{Baseline + $m$} demonstrates the effectiveness of the latent variable $m$. The latent variable not only identifies the POMDP-specific information and helps to learn POMDP-shared policy network, but also trades off the exploration and exploitation during the RL procedure. 3) The \textit{tran\_RS} and \textit{rew\_RS} are both effective because each of them encourages the policy learning stable. Compared to them, the superiority of  \textit{MetaVIM} indicates  \textit{tran\_RS} and \textit{rew\_RS} are complementary to each other. 4) Overall, all of the proposed components contribute positively to the final model.

\vspace{-3mm}
\section{Conclusions and Future Works}\label{sec:conclusions}
In this paper, we propose a novel Meta RL method MetaVIM for multi-intersection traffic signal control, which can make the policy learned from a training scenario generalizable to new unseen scenarios. MetaVIM learns the decentralized policy for each intersection which considers neighbor information in a latent way. We conduct extensive experiments and demonstrate the superior performance of our method over the state-of-the-art. We have collected and released more complex scenarios containing different structures \footnote{https://github.com/zhuliwen/RoadnetSZ}, and will improve the method based on these scenarios in the future. In addition, the  utilization of latent variable in model-based RL for traffic signal control will also be explored to improve sample efficiency.
\vspace{-3mm}

\ifCLASSOPTIONcaptionsoff
  \newpage
\fi




\bibliographystyle{IEEEtran}
\bibliography{bib}
%





%

\vspace{-1.2cm}
\begin{IEEEbiography}[{\includegraphics[width=1in,height=1.25in,clip,keepaspectratio]{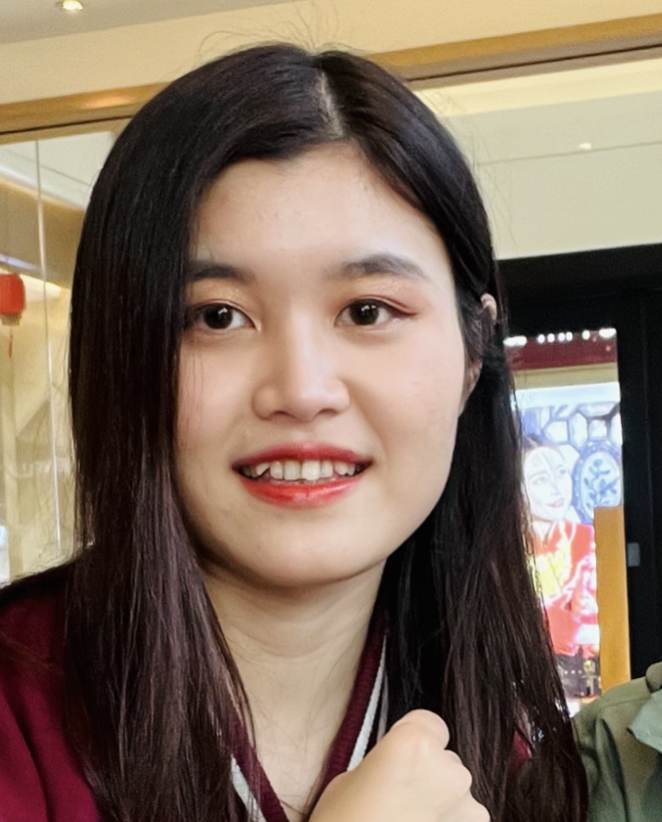}}]{Liwen Zhu} received the B.S. degree from Beijing University of Technology, China, in 2019. She is currently pursuing the M.S. degree with the School of Electronic and Computer Engineering, Peking University, China. Her current research interests include reinforcement learning and intelligent transportation system.
\end{IEEEbiography}
\vspace{-1.35cm}

\begin{IEEEbiography}[{\includegraphics[width=1in,height=1.25in,clip,keepaspectratio]{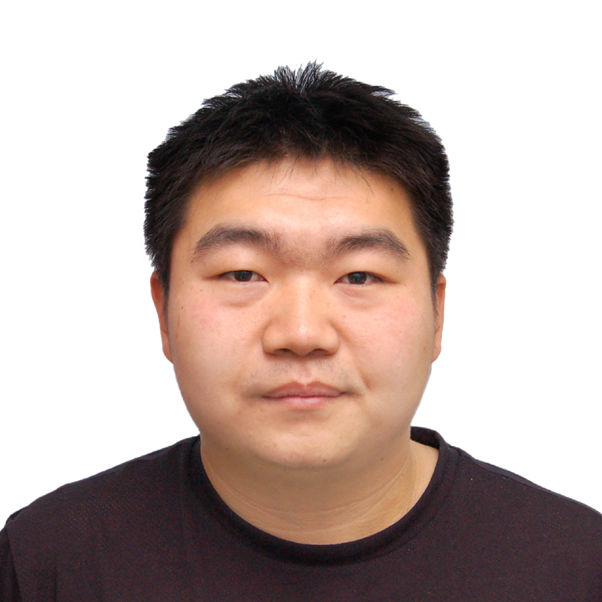}}]{Peixi Peng} received the PhD degree from  Peking University, in 2017, Beijing, China. He is currently an associate researcher with the School of Computer Science, Peking University, Beijing, China, and is also the assistant researcher of  Peng Cheng Laboratory, Shenzhen, China.  He is the author or co-author of more than 30 technical articles in refereed journals and top conferences.
His research interests include computer vision and reinforcement learning.
\end{IEEEbiography}
\vspace{-1.35cm}

\begin{IEEEbiography}[{\includegraphics[width=1in,height=1.25in,clip,keepaspectratio]{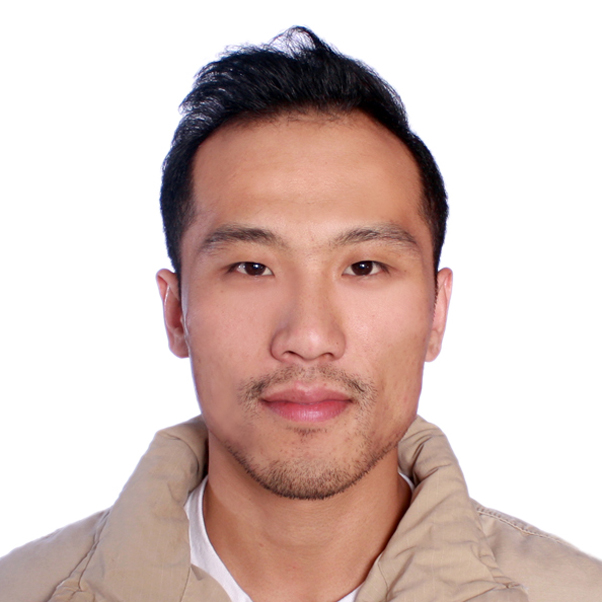}}]{Zongqing Lu} is a Boya Assistant Professor in the School of Computer Science, also affiliated with the Institute for Artificial Intelligence, Peking University.
He received the B.S. and M.S. degrees from Southeast University, China, and the Ph.D. degree from Nanyang Technological University, Singapore, 2014. His resarch interests include  (multi-agent) reinforcement learning and intelligent distributed systems.\end{IEEEbiography}
\vspace{-1.35cm}

\begin{IEEEbiography}[{\includegraphics[width=1in,height=1.25in,clip,keepaspectratio]{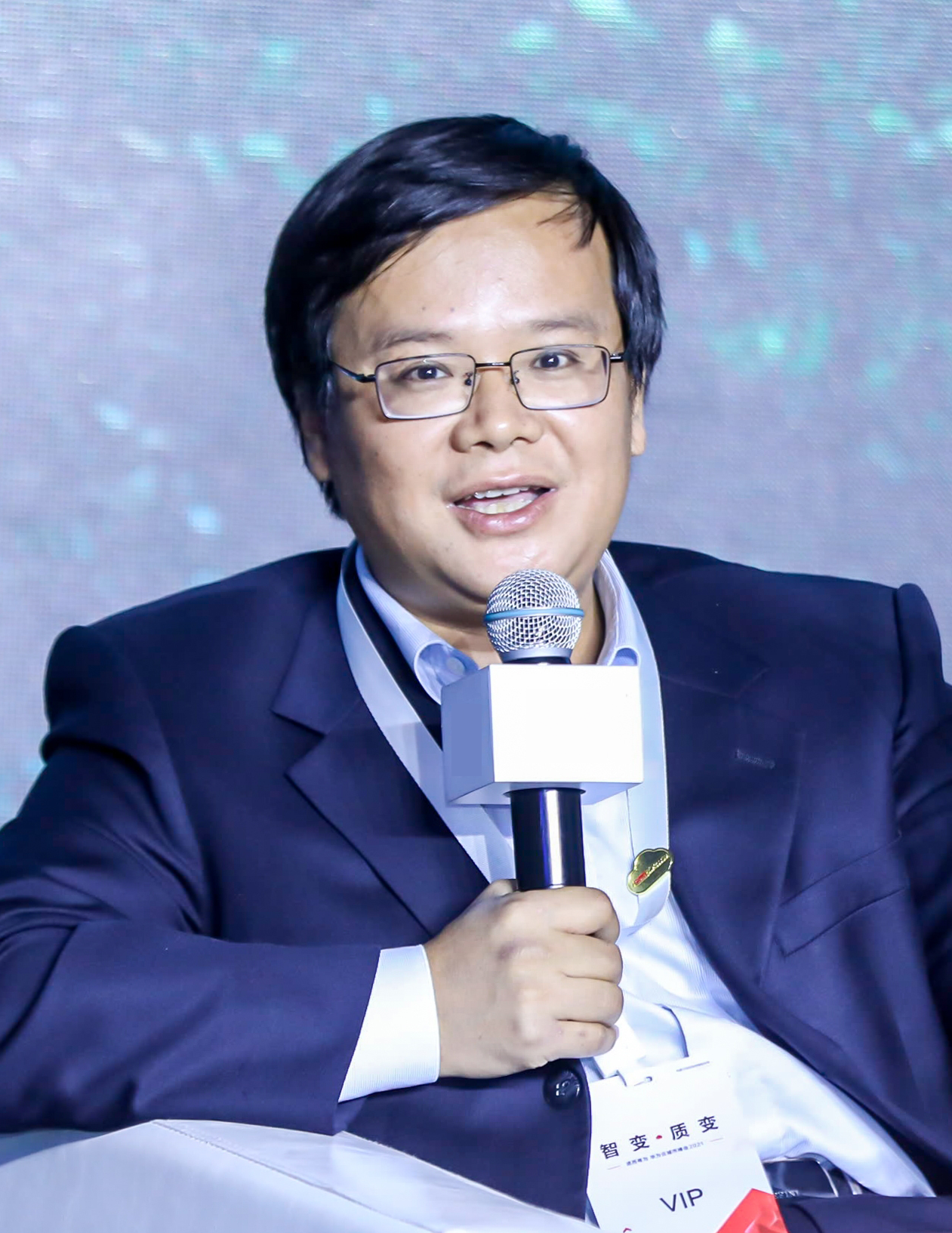}}]{Yonghong Tian}
is currently a Boya Distinguished Professor with the School of Computer Science, Peking University, China, and is also the deputy director of Artificial Intelligence Research Center, PengCheng Laboratory, Shenzhen, China. His research interests include neuromorphic vision, brain-inspired computation and multimedia big data. He is the author or coauthor of over 200 technical articles in refereed journals. He was the recipient of the Chinese National Science Foundation for Distinguished Young Scholars in 2018. He is a fellow of IEEE, a senior member of CIE and CCF, a member of ACM.
\end{IEEEbiography}

\vfill


\newpage
\onecolumn
\appendices

\section{Learning Curves}
\label{sec:learning_curves}
\begin{figure*}[h]
 		\centering
 		\includegraphics[width=0.98\textwidth]{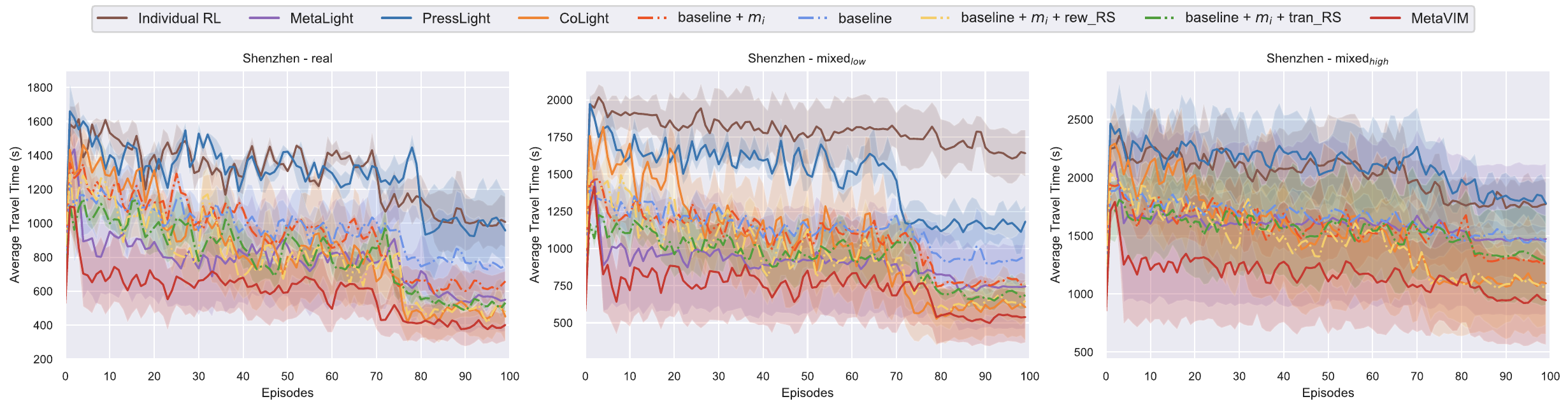} 
 		\vspace*{-0.2cm} 
 		\caption{Learning curves of MetaVIM and baselines in the Shenzhen road network map. The first, second and third columns correspond to the \textit{real}, \textit{mixed$_{l}$} and  \textit{mixed$_{h}$} configurations respectively. The final model MetaVIM outperforms all other models and shows great stability. In addition, the learning curves of ablations indicate that each component contributes positively.}
 		\label{fig:learning_curves_ablations}
\end{figure*}

\section{Implementation details of MetaVIM}
\label{sec:implementation_details}

\renewcommand\tabcolsep{5.5pt}  
\begin{table}[h]
    \footnotesize 
    \centering
    \setlength{\abovecaptionskip}{2pt}
    \caption{Implementation details of MetaVIM}
    \label{tab:implementation_details}
    \begin{tabular}{ll}
    \toprule 
    Items                           & Details                           \\
    \midrule
    Number of policy steps          & 3600                              \\ 
	Discount factor $\gamma$        & 0.95                              \\ 
	Policy minibatch                & 16                                \\
	mVAE minibatch                  & 25                                \\
	Value loss coefficient          & 0.5                               \\ 
	Entropy coefficient             & 0.01                              \\ 
	ELBO loss coefficient           & 1.0                               \\ 
	Latent space dimensionality     & 5                                 \\ 
	Aggregator hidden size          & 64                                \\
	\midrule
	Policy network                  & 2 hidden layers,                  \\
	architecture                    & 32 nodes each,                    \\
	                                & Tanh activations                  \\
	\midrule
	Policy network                  &  Adam with learning rate 0.0007   \\
	optimizer                       & and epsilon 1e-5                  \\
	\midrule
	                                & FC layer with 40 nodes,           \\
	                                & GRU with hidden size 64,          \\ 
	Encoder architecture            & output layer with                 \\
	                                & 10 outputs ($\mu$ and $\sigma$),  \\
	                                & ReLU activations                  \\
	\midrule 
	                                & 2 hidden layers,                  \\ 
	Transition Decoder              & 32 nodes each,                    \\ 
	architecture                    & 25 outputs heads,                 \\ 
	                                & ReLU activations                  \\
	\midrule
	                                & 2 hidden layers,                  \\ 
	Reward Decoder                  & 32 nodes each,                    \\ 
	architecture                    & 25 outputs heads,                 \\ 
	                                & ReLU activations                  \\
	\midrule
	mVAE optimizer                  &  Adam with learning rate 0.001    \\
	                                & and epsilon 1e-5                  \\
    \bottomrule
    \end{tabular}
\end{table}


\section{Experiments on different lane configurations}
\label{sec:experiments_on_different_lane_configurations}

\renewcommand\tabcolsep{16.0pt}  
\begin{table}[h]
    \normalsize 
    \centering
    \setlength{\abovecaptionskip}{3pt}
    \caption{Experimental performance on intersections with different topologies}
    \label{tab:multi_lane}
    \begin{tabular}{lrr}
    \toprule 
    \textbf{Model} & \textbf{l-2 / s-4 / r-1} & \textbf{l-1 / s-2 / r-1} \\
    \midrule
    Random & 828.00  & 866.33 \\
    MaxPressure & 829.06 & 867.16 \\
    Fixedtime & 1591.83  & 1529.09  \\
    FixedtimeOffset & 830.85  & 867.97  \\
    SOTL & 1258.67  & 1285.66  \\
    \midrule
    Individual RL & 835.64  & 860.37 \\
    MetaLight & 775.65 & 790.00 \\
    PressLight & 798.87 & 819.36  \\
    CoLight & 501.37 & 516.55  \\
    \midrule 
    MetaVIM & \textbf{485.36} & \textbf{490.12} \\ \bottomrule
    \vspace{-0.2cm}
    \end{tabular}
\end{table}

\normalsize
To validate that the method is robust to diverse lane configurations, we modify the 4x4 Hangzhou road network using different lane configurations:   \textit{left-2 / straight-4 /right-1} indicates the roads in the network is formed by 7 lanes, where the number of left-turn lanes, straight lanes and right-turn lanes are 1, 2, and 1 respectively. Similarly, \textit{left-1 / straight-2 /right-1} indicates the roads in the network is formed by 4 lanes, where the number of left-turn lanes, straight lanes and right-turn lanes are 1, 1, and 1 respectively. 

The results  are shown in Tab.~\ref{tab:multi_lane}, and it is evident that:
Among these baselines, the performance of Fixedtime is the worst because it can not adapt to the dynamics in the road network. RL-based methods show advantages than conventional traffic method. Among the RL-based method, MetaVIM outperforms Individual RL, MetaLight, PressLight and CoLight  in both \textit{left-2 / straight-4 /right-1} and \textit{left-1 / straight-2 /right-1} scenarios. It indicates the method can handle the intersections with different  lane configurations well.

\section{Experiments on SUMO}
\label{sec:experiments_on_sumo}

\renewcommand\tabcolsep{10.0pt}  
\begin{table*}[h]
    \normalsize 
    \centering
    \setlength{\abovecaptionskip}{3pt}
    \caption{Verification on the SUMO simulation platform}
    \label{tab:sumo_exp}
    \begin{tabular}{l l l l l l}
    \toprule 
    Metrics & MA2C & IA2C & IQL-LR & IQL-DNN & MetaVIM \\
    \midrule
    reward & -366.13 & -1062.23 & -1057.19 & -2619.42 & -298.58 \\
    avg. queue length [veh] & 1.88 & 3.74 & 2.78 & 4.48 & 3.97 \\
    avg. intersection delay [s/veh] & 11.98 & 52.52 & 74.56 & 166.03 & 10.56 \\
    avg. vehicle speed [m/s] & 2.40 & 1.60 & 3.18 & 1.48 & 2.86 \\
    trip completion flow [veh/s] & 0.63 & 0.38 & 0.73 & 0.23 & 0.79 \\
    trip delay [s] & 322.81 & 560.61 & 186.67 & 477.43 & 157.23 \\
    \bottomrule
    \end{tabular}
\end{table*}

To validate that the method is robust to different traffic simulation platforms, several experiments are conducted on  SUMO\footnote{http://sumo.dlr.de/index.html
}. For fair comparisons, 4 recent methods which provides the source code on SUMO are evaluated at the same setting, including:

$\bullet$ MA2C \cite{chu2019multi}: a fully scalable and decentralized MARL alogrithm for the deep RL agent: advantage actor critic (A2C), within the context of adaptive traffic signal control (ATSC).

$\bullet$ IA2C \cite{chu2019multi}: an independent advantage actor critic (A2C) method in multi-agent scenario.

$\bullet$ IQL-LR \cite{cai2009adaptive}: a control algorithm based on dynamic programming, using an approximation ot the value function of the dynamic programming and reinforcement learning to update the approximation.

$\bullet$ IQL-DNN \cite{chu2019multi}: the same as IQL-LR but uses DNN for fitting the Q-function.

We choose a 5x5 large traffic grid as the experimental scenario, which is the same as \cite{chu2019multi}. The scenario is formed by two-lane arterial streets with speed limit 20m/s and one-lane avenues with speed limit 11m/s. The evaluation metrics are provided by SUMO as follows:

$\bullet$ \textit{Reward}
The rewards received by all agents across the entire simulation time.

$\bullet$ \textit{Avg. Queue Length}
Average queue length is the average numbers of vehicles in the queue over all intersections. If a vehicle is waiting at an intersection and the speed less than 0.1m/s, we think the car is on the queue.

$\bullet$ \textit{Avg. intersection delay}
If a vehicle is waiting at an intersection and the speed less than 0.1m/s, the extra time the vehicle stays at the intersection is the delay time. We calculate the average time over all vehicles, then we get the avg. intersection delay.

$\bullet$ \textit{Avg. vehicle speed}
The average speed of all vehicles in the road network for the entire simulation time.

$\bullet$ \textit{Trip completion flow}
Trip completion flow is the number of vehicles that complete the trip. Trip completion flow is calculated by dividing the total number of vehicles that complete the trip during the entire simulation time by the horizon.

$\bullet$ \textit{Trip delay}
Trip delay is the extra time wasted for all vehicles that complete the trip in the road network.

Tab. \ref{tab:sumo_exp} lists the comparative results on the common testing mode over SUMO, and it is evident that:
In general, MetaVIM performs better than advantage actor critic and Q learning in both decentralized MARL and independent control, and it indicates the advantage of the MetaVIM in various evaluation criterias. MetaVIM achieves higher reward in evaluation. Moreover, MetaVIM is superior to other methods clearly in average queue length, average intersection delay, average vehicle speed, trip completion flow and trip delay, which demonstrates the effectiveness of the method in SUMO platform. 

\section{Scalability Validation}
\label{sec:scalability_validation}
\begin{figure}[h]
	\centering
	\vspace{-0.2cm}
	\includegraphics[width=0.26\textwidth]{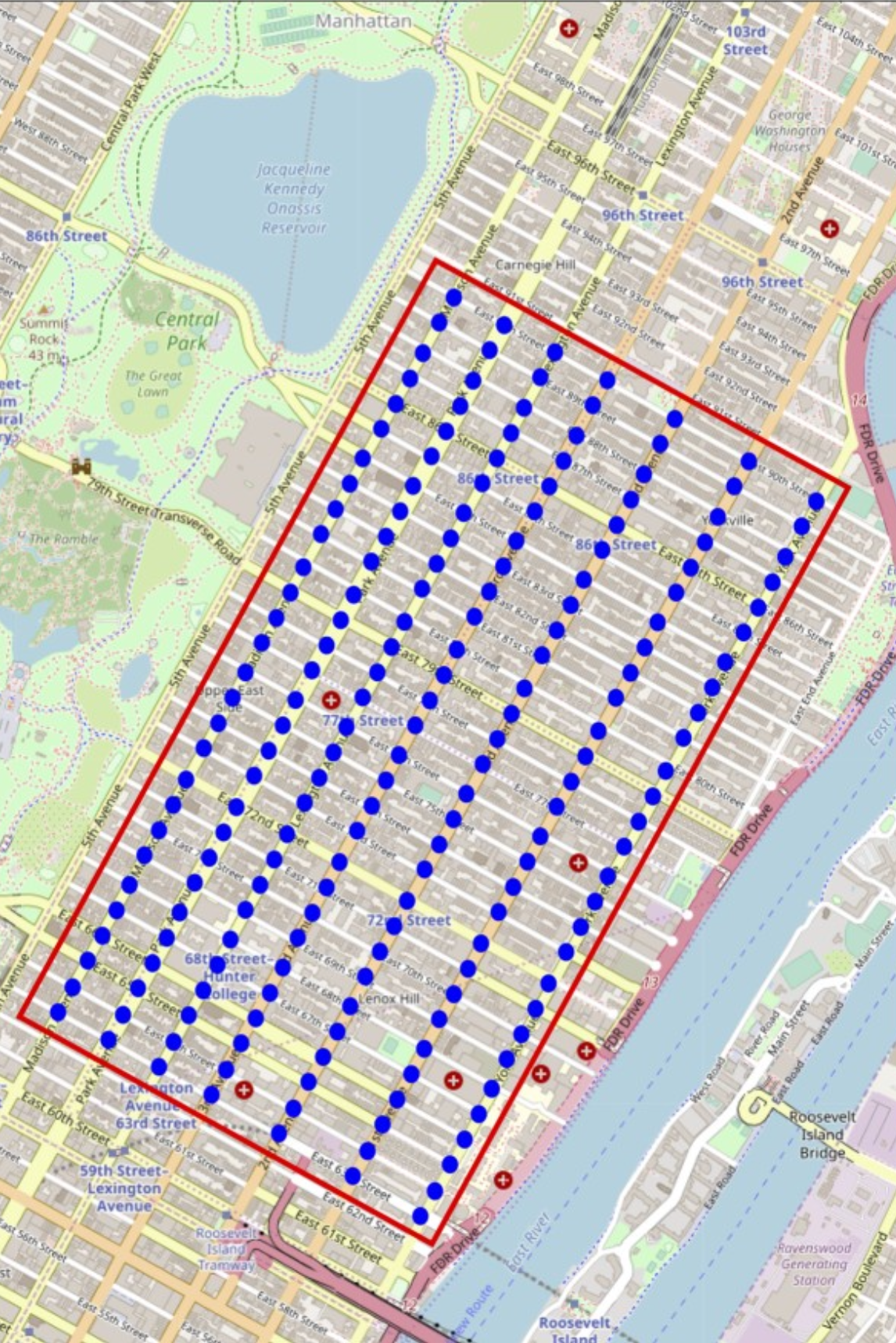}
	\caption{Simulation on a large-scale dataset with 196 intersections}
	\vspace{-0.4cm}
	\label{fig:newyork196}
\end{figure}

To validate scalability of the method, an additional experiment is conducted on a public dataset of 196 intersections in New York City \footnote{\url{https://github.com/wingsweihua/colight/blob/master/data/NewYork/28_7/roadnet_28_7.json}}, as illustrated in Figure \ref{fig:newyork196}. It is one of the largest public dataset in Cityflow to our knowledge. As shown in  Tab. \ref{tab:newyork196}, MetaVIM achieves better results than compared methods, which demonstrates the superior scalability of MetaVIM. The reason is that  MetaVIM is a decentralized method and doesn't need the joint action of all agents and full state. Hence, MetaVIM could avoid the dimensional explosion of large scale of agents. In addition, as the number of agents increases, more sample data will be collected. It leads to a significant increase in the number of training samples, rather than an increase in dimensionality. Therefore,  MetaVIM could scale well.

\renewcommand\tabcolsep{15.0pt}
\begin{table}[h]
    \normalsize 
    \centering
    \setlength{\abovecaptionskip}{3pt}
    \caption{Experimental performance on 196 intersections in New York}
    \label{tab:newyork196}
    \begin{tabular}{lrrrr}
    \toprule 
    \textbf{Model} & \textbf{real} & \textbf{mixed$_l$} & \textbf{mixed$_h$} & mean \\
    \midrule
    Random & 1869.00  & 1748.76 & 1892.73 & 1836.83 \\
    MaxPressure & 845.72 & 589.24 & 920.31 & 785.09 \\
    Fixedtime  &  1831.65 & 1742.71  & 1205.97 & 1593.44 \\
    FixedtimeOffset & 1702.82 & 1732.37 & 1869.66 & 1768.28 \\
	SlidingFormula & 1185.64 & 921.28 & 1299.19 & 1135.37 \\
    SOTL & 1862.34  & 1793.37  & 1939.53 & 1865.08 \\
    \midrule
    Individual RL & 1877.46 & 1369.64 & 1906.43 & 1717.84 \\
    MetaLight & 1285.74 & 1005.36 & 1368.96 & 1220.02 \\
    PressLight & 1586.75 & 1547.74 & 1685.74 & 1606.74 \\
    CoLight & 637.79 &  479.31 & 761.10 & 626.07 \\
	GeneraLight & 710.23 &  485.84 & 862.47 & 686.18 \\
    \midrule 
    MetaVIM & \textbf{586.36} & \textbf{396.57} & \textbf{748.65} & \textbf{577.19} \\ \bottomrule
    \end{tabular}
\end{table}

\section{RL-based Traffic Signal Control Survey}
\label{sec:rl_based_tsc}

A survey on RL-based traffic signal control methods is shown in Tab. \ref{tab:related_work}.

\renewcommand\arraystretch{1.47}
\renewcommand\tabcolsep{8.0pt}
\begin{table}[h]
    \small
    \centering
    \caption{RL-based Methods Summary}
    \label{tab:related_work}
    \begin{tabular}{m{1.8cm}<{\centering}|m{3.4cm}|m{10.5cm}}
    \toprule 
    \textbf{Classes} & \textbf{Methods} & \textbf{Descriptions}\\
    \midrule
							& MARLIN-ATSC \cite{el2013multiagent} & real-time adjustment of signal timing plans based on traffic fluctuations, with each agent coordinating with adjacent intersections to generate actions  \\ 
		tabular Q-learning	& holonic Q-learning \cite{abdoos2013holonic} & the traffic network is divided into multiple zones with two phases of control  \\
                            & DWL \cite{dusparic2009distributed} & collaboration-based agent self-optimizes for multiple strategies  \\
							& Q-learning based \cite{abdoos2011traffic} & apply traditional Q-learning method to multi-agent systems   \\
    \midrule
                            & Intellilight \cite{wei2018intellilight} & use a variety of different combinations of traffic indicators as the states \\   
                            & RLTSC \cite{mannion2016experimental} & use current phase index, phase duration, and queue length as states \\
        single agent        & AttendLight \cite{oroojlooy2020attendlight} & a single general model is designed, and two attention models are used to deal with multiple topologies \\
                            & MT-GAD \cite{jiang2021dynamic} & use a group attention structure to reduce the number of required parameters and to achieve a better generalizability \\
                            & PG Time EWMA \cite{rizzo2019time} & a policy gradient method based on episodic conditions and a time-dependent baseline to learn an optimal policy for traffic signal control in congested conditions \\
    \midrule
                            & LIT \cite{zheng2019diagnosing} & analyze which traffic indicators are suitable as states or reward \\
                            & FRAP \cite{zheng2019learning} & introduce phase competition to select a more suitable phase \\
    isolated multi-agent    & DemoLight \cite{xiong2019learning} & demonstrations can be collected from classical methods to accelerate learning \\
                            & PressLight \cite{wei2019presslight} & combine the traditional traffic method MaxPressure with RL technology together \\
                            & MPLight \cite{chen2020toward} & experiment on a large dataset using a shared strategy \\
                            & PlanLight \cite{zhang2020planlight} & learn from the demonstration of the rollouts \\
    \midrule
    centralized             & DQN-TP \cite{van2016coordinated} & combination of DQN method and transfer algorithm to overcome instability problem \\
                            & max-plus  \cite{kuyer2008multiagent} & coordination by max-plus algorithm to obtain optimal joint actions \\
    \midrule
                            & LQF \cite{arel2010reinforcement} & a distributed static control approach that integrates adaptive RL systems and LQF \\
                            & NFQI with GCNN  \cite{nishi2018traffic} & the traffic characteristics of long-distance roads are considered by using graph neural network \\
    decentralized           & CoLight \cite{wei2019colight} & use graph convolution and attention mechanism to model the neighbor information \\
                            & HiLight \cite{xu2021hierarchically} & use hierarchical reinforcement learning to allow agents to optimize different short-term sub-goals \\
                            & MA2C \cite{chu2019multi} & use neighbor information to improve observability and reduce the learning difficulty of local agents \\
    \midrule
                            & MetaLight \cite{zang2020metalight} & a value-based meta reinforcement learning method via parameter initialization \\
    meta-learning           & GeneraLight \cite{zhang2020generalight} & generate diversified traffic flows using GAN, and then improve generalization ability through clustering \\
                            & ModelLight \cite{huang2021modellight} & a model-based meta-reinforcement learning framework, meta-learning method can improve data efficiency and reduce the number of interactions required with real-world environments \\
\bottomrule
    \end{tabular}
\end{table}

\end{document}